\documentclass[12pt]{article}
\pdfoutput=1

\usepackage[preprint]{n_edited}
\usepackage{setspace} 
\usepackage{hyperref}
\usepackage[dvipsnames,table,xcdraw]{xcolor}
\usepackage[most]{tcolorbox}
\usepackage{pdfpages}
\newtcolorbox{mybox}
{
  enhanced jigsaw,
  breakable,
  pad at break*=1mm,
  colback=blue!4!white,
  colframe=blue!75!black
}

\hypersetup{
  linkcolor  = violet!85!black,
  citecolor  = YellowOrange!85!black,
  urlcolor   = Aquamarine!85!black,
  colorlinks = true,
}

\hypersetup{breaklinks=true}
\usepackage{xurl}
\usepackage{enumitem}
\setlist{leftmargin=5mm}

\usepackage{float} % to force the position of some figures
\usepackage{wrapfig}

\PassOptionsToPackage{hyphens}{url}
\usepackage{booktabs}       % professional-quality tables
\usepackage{amsfonts}       % blackboard math symbols
\usepackage{nicefrac}       % compact symbols for 1/2, etc.
\usepackage{microtype}      % microtypography
\usepackage[dvipsnames,table,xcdraw]{xcolor}         % colors
\usepackage{multirow}
\usepackage{nicematrix}
\usepackage[labelfont=bf]{caption}

\usepackage{hhline}
\usepackage{cellspace}
\setcitestyle{numbers}
\def\mybibitem{%
  \vskip0pt%\baselineskip% % space above
  \noindent%           % suppress regular indent
  \hangindent=2.8em%   % indent same amount as `quotation` environment
}

\title{International Governance of Civilian AI:\\ 
A Jurisdictional Certification Approach}

\author{Robert F. Trager,$^{1,2}$ 
Ben Harack,$^3$
Anka Reuel,$^4$
Allison Carnegie,$^5$\\
\textbf{Lennart Heim,$^2$}
\textbf{Lewis Ho,}
\textbf{Sarah Kreps,$^{6,7}$}
\textbf{Ranjit Lall,$^3$}
\textbf{Owen Larter,$^8$}\\ 
\textbf{Seán Ó hÉigeartaigh,$^9$}
\textbf{Simon Staffell,$^8$}
\textbf{José Jaime Villalobos$^{10}$}
\vspace{10pt}\\
 $^1$Blavatnik School of Government, University of Oxford, $^2$Centre for the Governance of AI,\\ $^3$University of Oxford,
 $^4$Stanford University, 
 $^5$Columbia University,
 $^6$Tech Policy Institute,\\ Cornell University,
 $^7$Brookings Institution, $^8$Microsoft,
 $^9$Centre for the Future of Intelligence,\\ University of Cambridge,
 $^{10}$Legal Priorities Project}

\usepackage[T1]{fontenc}
\usepackage{fourier}
\usepackage[scaled=1]{Baskervaldx}
\begin{document}
%\includepdf{images/titlepage}
\setcounter{page}{1}
\maketitle

\vspace{.5cm}

\vspace{-\baselineskip}\centerline{\emph{This report does not necessarily represent the views of the co-authors’ employers.}}
\medskip

\vspace{.5cm}

\begin{abstract}
This report describes trade-offs in the design of international governance arrangements for civilian artificial intelligence (AI) and presents one approach in detail. This approach represents the extension of a standards, licensing, and liability regime to the global level. We propose that states establish an International AI Organization (IAIO) to certify state \emph{jurisdictions} (not firms or AI projects) for compliance with international oversight standards. States can give force to these international standards by adopting regulations prohibiting the import of goods whose supply chains embody AI from non-IAIO-certified jurisdictions. This borrows attributes from models of existing international organizations, such as the International Civil Aviation Organization (ICAO), the International Maritime Organization (IMO), and the Financial Action Task Force (FATF). States can also adopt multilateral controls on the export of AI product inputs, such as specialized hardware, to non-certified jurisdictions. Indeed, both the import and export standards could be required for certification. As international actors reach consensus on risks of and minimum standards for advanced AI, a jurisdictional certification regime could mitigate a broad range of potential harms, including threats to public safety. 
\end{abstract}

{\footnotesize \noindent \textit{Corresponding authors}: Robert Trager (robert.trager@governance.ai), Ben Harack (ben.harack@gmail.com), Anka Reuel (anka@cs.stanford.edu)}

\newpage

\section*{Executive Summary}\addcontentsline{toc}{section}{Executive Summary}

\vspace{1cm}

\subsection*{AI Risks Require International Governance}

As automated systems of unprecedented capabilities are developed and deployed, society faces an extraordinary governance challenge, with new risks ranging from algorithmic bias to threats to public safety. Domestic regulation is being developed in states with leading AI capabilities, but domestic regulation is not sufficient. 

The AI industry is deeply international, with supply and product networks spanning many states. While research efforts in a few states are at the forefront, technological understanding of the essential elements of creating frontier systems is becoming more widely dispersed. Though specialized AI-chip supply chains are highly concentrated, access to computing resources is also dispersed. The relatively low computing requirements for using current systems—as opposed to building them—mean that even the most advanced systems can be used by many firms and states around the world that gain access to the trained models. The evolving geographic distribution of AI capabilities is thus uncertain, with global inequities in need of correction, but it is likely that actors in many states will gain access to capabilities sufficient to pose risks of societal harms.

The potential harms of AI can also cross state borders. Many AI models are accessible online via either API access or an open-source version, which contributes to an immediate global impact. In the future, proprietary systems might be copied against the wishes of their creators. Biological and chemical agents designed by AI technologies could be released far from where they are designed. AI-enabled propaganda or spear phishing campaigns can target people in any country. Competition among firms and states can pressure them into taking greater risks with the technology. These risks and interactions may culminate in catastrophic risks. Indeed, dozens of leading AI scientists have signed a statement that “mitigating the risk of extinction from AI should be a global priority”.\footnote{Center for AI Safety, “Statement on AI Risk,” May 30, 2023, \url{https://www.safe.ai/statement-on-ai-risk}.} 

Thus, international governance of AI is a global problem in which all have a stake, but presently not all have meaningful representation and input. Regulating AI on a country-by-country basis will likely lead to inadequate regulation in some jurisdictions and fragmented and disjointed regulation in others, hampering needed international collaboration on AI safety and global development. Taking into account the particular characteristics of the AI industry, this report describes trade-offs in the design of international governance of AI and presents one approach to \emph{civilian AI}\footnote{Civilian AI refers to all AI \emph{except} that built under the direct authority of the state for sensitive purposes such as the military or intelligence services.} governance in detail. We focus on regulating \emph{frontier AI}, though the approach could be applied more broadly. We define frontier AI as models “trained on broad data at scale in order to be generally useful across tasks” (i.e. “foundation models”) with capabilities sufficient to pose risks to public safety.\footnote{Rishi Bommasani et al., “On the Opportunities and Risks of Foundation Models” (arXiv, 2022), \href{https://doi.org/10.48550/arXiv.2108.07258}{arXiv:2108.07258}; Markus Anderljung et al., “Frontier AI Regulation: Managing Emerging Risks to Public Safety” (arXiv, 2023), \href{https://doi.org/10.48550/arXiv.2307.03718}{arXiv:2307.03718}.}

\subsection*{Civilian Frontier AI Governance Is Urgently Needed and Feasible}

With most frontier AI development occurring in the private sector, regulating civilian AI is a policy priority, and we expect the serious risks from AI to arise initially in that domain. The international civilian governance problem is urgent because some of the risks described above are present already, and system capabilities and their associated risks are expected to grow rapidly as systems scale and algorithms improve due to large investments in the sector. Military arms control is also desirable, but progress there will be relatively slow and challenging. By contrast, existing models of international civilian regulation appear applicable to civilian frontier AI and compatible with states’ interests. Moreover, efforts to govern these different domains can be synergistic, since civilian AI governance can provide a useful testing ground for processes and mechanisms that might eventually be used for governing militaries.

\subsection*{A Proposal for an International Governance System for Civilian AI}

We propose a set of international institutions that allow for civilian AI regulations to be consistently applied across jurisdictions—when sufficient international consensus exists on minimum regulatory standards. States can coordinate to create an International AI Organization (IAIO) to certify state \emph{jurisdictions} for compliance with international oversight standards. States can give force to these international standards by adopting regulations prohibiting the import of goods whose supply chains integrate AI from non-IAIO-certified jurisdictions. Further weight can be given to these standards if states adopt controls on the export of AI product inputs, such as specialized chips, to non-certified jurisdictions. This approach borrows attributes from the pattern of existing international organizations, such as the International Civil Aviation Organization (ICAO), the International Maritime Organization (IMO), and the Financial Action Task Force (FATF).

\subsection*{Incentives for Participation}

The proposed structure benefits all states, including both technology leaders and developing states. All states can protect themselves from the harms of AI while retaining access to an international market with consistent regulations. States with cutting-edge AI industries can design their regulatory agencies to minimize proliferation of industry secrets. Developing states can participate at low cost, especially if the IAIO provides direct firm-monitoring capacity as a service to states that want it—thus allowing all states to benefit from the pooling of monitoring capacity. The IAIO or a separate organization should also be tasked with the international sharing of safe AI technologies and facilitating broad access to the benefits of the technology.

\subsection*{Enforcement and Robustness}

Enforcement of the regime would be via conditional market access: requiring certification in order to freely trade AI precursors and products. Concretely, this enforcement would be enacted via domestic laws in each participating state. One option to increase the strength of enforcement is to require—after some lead time—that states embed enforcement provisions in their laws as a condition of IAIO certification. Such a mechanism reverses the typical collective-action problem of international enforcement, since collective action would be required if states wanted to \emph{avoid} enforcing the regime. The threat of being cut off from AI markets in participating states provides all states with an incentive to join the regime and stay in compliance.

\begin{table}[ht!]
    \centering
    \tcbincludegraphics[arc=0mm,left=0mm, right=0mm, top=0mm, bottom=0mm, colback=white,width=\linewidth,graphics options={trim=8mm 8mm 8mm 15mm, clip}]{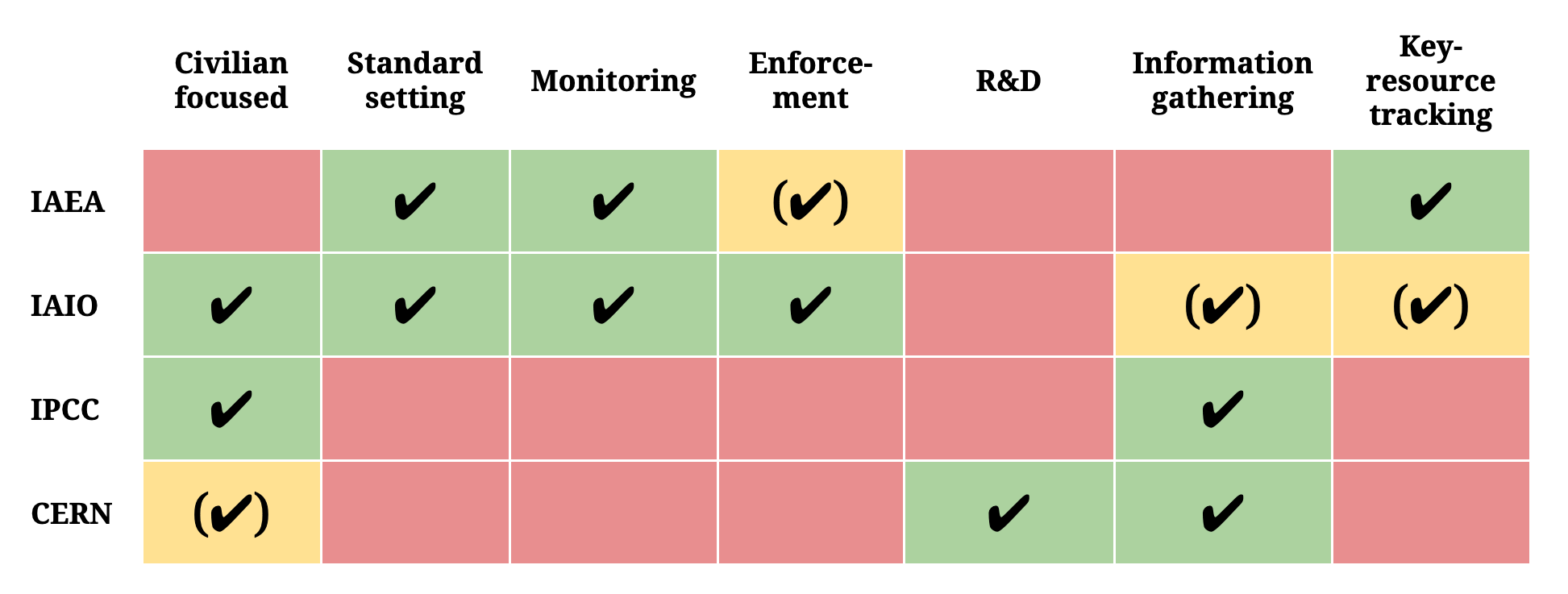}   \setstretch{0.85} 
    \vspace{-3pt}
    \parbox{\linewidth}{\begin{footnotesize}\emph{Note}: Green indicates that the model fulfills this function; red indicates that it does not. Yellow means that there is some ambiguity; for instance, the IAEA only refers violations to the Security Council which then potentially takes action, a process that could be counted as enforcement. Similarly, tracking of key AI inputs could be part of the IAIO model but is optional. In the case of CERN, despite its civilian focus, the research could be classified as dual-use to a degree. These institutions were chosen for comparison because they represent commonly discussed models for international AI governance.\protect\footnotemark\ The IAIO is based on the ICAO, IMO, and FATF models, and thus these are not listed because they share similar characteristics.\end{footnotesize}\\}
    \caption{\textbf{Features of institutional analogies for AI governance models.}}     
    \label{table1}
\end{table}
\footnotetext{See Lewis Ho et al., “International Institutions for Advanced AI” (arXiv, 2023), \href{https://doi.org/10.48550/arXiv.2307.04699}{arXiv:2307.04699}.}

\subsection*{Contrast to Other Approaches to International AI Governance}

We highlight one approach to an international regime for civilian AI standard setting, monitoring, and enforcement, but other approaches to international governance should also be considered. Table~\ref{table1} summarizes some of the key differences of this approach from other proposals. The IAIO model enables agile standard setting, monitoring, and enforcement by focusing on internationally agreed-upon minimum safety standards for the global industry, international jurisdictional monitoring, and state enforcement. One difference between the proposed IAIO and an institution modeled after the International Atomic Energy Agency (IAEA) is that domestic regulators, rather than an international organization, would implement standards and interact with local firms, easing proliferation concerns of states with frontier labs and enabling rapid responses to standards violations. 

\subsection*{Recommendations}

\begin{itemize}[itemsep=-11pt]        
    \item Develop consensus on minimum regulatory standards and model evaluations for civilian AI through continued dialog with national regulators, civil society, academia, industry, and international organizations such as the United Nations, OECD and others.\\
    \item Encourage and support states in creating domestic regulatory capacities for AI.\\
    \item Use a global summit to agree on milestones for setting up an international civilian AI regulatory regime resembling the existing standards harmonization regimes centered on the ICAO, IMO, and FATF.
    \begin{itemize}
        \item Agreement on the structure of the regime, as distinct from developing the standards themselves, should be designed to complete within six months of the summit. 
        \item Milestones should include agreement on the types of risks the regime would focus on and the core elements and principles of the proposed organization’s functioning, such as the process for creating standards and the nature of the interaction between the proposed organization and domestic regulators.
        \item A core group of experts and representatives from both frontier and non-frontier states can manage the process with input from all UN states as well as non-governmental stakeholders, such as relevant NGOs, unions, and consumer groups.
        \item In parallel, actors should consider initiating the governance regime among smaller sets of actors with the intention of expanding to include other actors over time. Starting with a small set of actors may be necessary for near-term agreement and, as an outside option, may facilitate agreement between a broad set of actors.
        \item The board and decision-making procedures of the proposed international organization should be structured to respect the interests of both frontier and non-frontier states and mitigate against the organization being employed for political ends outside of its mandate. The board should contain representatives from the technical and civil society AI governance communities, frontier AI states, and non-frontier AI states. 
        \item Special care will be needed to prevent states from attempting to use a monitoring organization to gain access to frontier lab technologies.\\
    \end{itemize}
    \item Explore an AI-specialized-computing-hardware ownership registry with unique hardware IDs to enable future governance efforts that benefit from computing-capacity transparency.
\end{itemize}

\clearpage

\tableofcontents

\clearpage 

\section{Introduction}

Artificial intelligence (AI)\footnote{Jonas Schuett, ``Defining the Scope of AI Regulations,'' Legal Priorities Project Working Paper No. 9 (2021), \url{https://papers.ssrn.com/abstract=3453632}.} systems are having ever greater impacts on societies, leading to calls for international governance. In recent months, the importance of international AI governance has been noted by politicians and industry leaders in meetings at the White House and a US Senate committee;\footnote{The White House, ``Readout of White House Meeting with CEOs on Advancing Responsible Artificial Intelligence Innovation,'' May 4, 2023, \url{https://www.whitehouse.gov/briefing-room/statements-releases/2023/05/04/readout-of-white-house-meeting-with-ceos-on-advancing-responsible-artificial-intelligence-innovation/}.}\footnote{``Oversight of A.I.: Rules for Artificial Intelligence,'' May 16, 2023, \url{https://www.judiciary.senate.gov/committee-activity/hearings/oversight-of-ai-rules-for-artificial-intelligence}.} statements by the United Nations Secretary-General,\footnote{António Guterres, ``Secretary-General’s Remarks to the Security Council on Artificial Intelligence'' (United Nations Security Council, July 18, 2023), \url{https://www.un.org/sg/en/content/sg/speeches/2023-07-18/secretary-generals-remarks-the-security-council-artificial-intelligence}.} BRICS nations,\footnote{ABP News Bureau, ``BRICS Nations Call For Effective Global Framework On AI, Emphasise On Ethical Development,'' ABP News Live, June 2, 2023, \url{https://news.abplive.com/technology/ai-brics-nations-call-for-effective-global-framework-on-artificial-intelligence-emphasise-on-ethical-development-1606406}.} OpenAI,\footnote{Sam Altman, Greg Brockman, and Ilya Sutskever, ``Governance of Superintelligence,'' OpenAI, May 22, 2023, \url{https://openai.com/blog/governance-of-superintelligence}.} Google DeepMind,\footnote{\emph{Google CEO Calls for Global AI Regulation} (60 Minutes, April 16, 2023), \url{https://www.youtube.com/watch?v=aNsmr-tvQhA}.} and Microsoft;\footnote{Brad Smith, ``Governing AI: A Blueprint for the Future'' (Microsoft, May 25, 2023).} leaders of the UK and US pledging to work together on AI safety;\footnote{``Britain, U.S. to Work Together on AI Safety, Says Sunak,'' Reuters, June 8, 2023, \url{https://www.reuters.com/technology/britain-us-work-together-ai-safety-says-sunak-2023-06-08/}.} and plans for a Global Summit on AI Safety.\footnote{``UK to Host First Global Summit on Artificial Intelligence,'' GOV.UK, June 7, 2023, \url{https://www.gov.uk/government/news/uk-to-host-first-global-summit-on-artificial-intelligence}.} Yet, even those calling for international governance appear to have only nascent ideas about what sorts of governance would be feasible and would achieve the best global economic and security outcomes. 

In this report, we identify the landscape of approaches to international \emph{civilian} AI governance and describe one approach in detail.\footnote{We define ``civilian'' AI as all AI except that built under the direct authority of the state for sensitive purposes such as the military or intelligence services.} We contend that international civilian and state/military AI, where the distinction is based on the application context of the technology, should have separate governance processes because they differ in key ways that shape how they can be governed. Throughout, we focus on regulating \emph{frontier AI}, though the approach we recommend could be applied more broadly. We define frontier AI as models ``trained on broad data at scale in order to be generally useful across tasks'' (i.e. ``foundation models'') with capabilities sufficient to pose significant risks to public safety.\footnote{Bommasani et al., ``On the Opportunities and Risks of Foundation Models.''; Anderljung et al., ``Frontier AI Regulation.''} Building on an overview of the international AI industry, we explain why international governance of the AI ecosystem is needed. We describe important trade-offs in the design of international institutions relating to standards, monitoring, enforcement, and institutional governance. Finally, we discuss a promising approach to civilian AI governance and its associated benefits and challenges.

The approach we describe extends a standards, licensing, and liability regime to the global level. We propose that states coordinate to create an International AI Organization (IAIO) to certify state jurisdictions (not firms or AI projects) for compliance with international oversight standards. States can enforce these international standards by adopting regulations prohibiting the import of goods whose supply chains embody AI from non-IAIO-certified jurisdictions. This follows the models of some existing international organizations, such as the International Civil Aviation Organization (ICAO), the International Maritime Organization (IMO), and the Financial Action Task Force (FATF). States can also adopt controls on the export of AI product inputs, such as specialized chips, to non-certified jurisdictions. Indeed, these import and export standards could be required for certification. We describe how such a regime would have wide applicability in mitigating many of advanced AI’s potential harms, from algorithmic bias to threats to public safety.

\section{Scope of the AI Governance Challenge}

Existing AI systems are capable of extraordinary things, and progress has been rapid. Among other remarkable feats, AI-based systems have contributed to solving key scientific puzzles,\footnote{John Jumper et al., ``Highly Accurate Protein Structure Prediction with AlphaFold,'' \emph{Nature} 596, no. 7873 (2021): 583–89, \url{https://doi.org/10.1038/s41586-021-03819-2}.} passed informal versions of the Turing Test (once believed to be the most important test of human-level intelligence),\footnote{Daniel Jannai et al., ``Human or Not? A Gamified Approach to the Turing Test'' (arXiv, 2023), \href{https://doi.org/10.48550/arXiv.2305.20010}{arXiv:2305.20010}; although see also Sharon Temtsin, Diane Proudfoot, and Christoph Bartneck, ``A Bona Fide Turing Test,'' in \emph{Proceedings of the 10th International Conference on Human-Agent Interaction}, HAI ’22 (ACM, 2022), 250–52, \url{https://doi.org/10.1145/3527188.3563918}.} become the fastest-growing product in history,\footnote{Andrew Chow, ``How ChatGPT Managed to Grow Faster Than TikTok or Instagram,'' \emph{Time}, February 8, 2023, \url{https://time.com/6253615/chatgpt-fastest-growing/}.} and scored well on the uniform bar exam.\footnote{OpenAI, ``GPT-4 Technical Report'' (arXiv, 2023), \url{https://doi.org/10.48550/arXiv.2303.08774}{arXiv:2303.08774}; challenged by Eric Martínez, ``Re-Evaluating GPT-4’s Bar Exam Performance'' (SSRN, 2023), \url{https://doi.org/10.2139/ssrn.4441311}.} Along the way, leading AI systems have also demonstrated surprisingly \emph{general} capabilities, where for example a single model can perform well on a broad set of tasks, including understanding and generating text in many languages, scoring well on standardized tests, and writing computer code.\footnote{OpenAI, ``GPT-4 Technical Report.''} 

Recent progress in AI has been undergirded by improved AI algorithms,\footnote{Ege Erdil and Tamay Besiroglu, ``Algorithmic Progress in Computer Vision'' (arXiv, 2023), \href{https://doi.org/10.48550/arXiv.2212.05153}{arXiv:2212.05153}; Jordan Hoffmann et al., ``Training Compute-Optimal Large Language Models'' (arXiv, 2022), \href{https://doi.org/10.48550/arXiv.2203.15556}{arXiv:2203.15556}.} increased investment,\footnote{Ben Cottier, ``Trends in the Dollar Training Cost of Machine Learning Systems'' (Epoch, 2023), \url{https://epochai.org/blog/trends-in-the-dollar-training-cost-of-machine-learning-systems}.} and dramatic increases in spending on compute—the computational hardware used to train models.\footnote{Saif M. Khan and Alexander Mann, ``AI Chips: What They Are and Why They Matter'' (Center for Security and Emerging Technology, 2020), \url{https://doi.org/10.51593/20190014}; Jaime Sevilla et al., ``Compute Trends Across Three Eras of Machine Learning'' (arXiv, 2022), \href{https://doi.org/10.48550/arXiv.2202.05924}{arXiv:2202.05924}.} All of these trends are expected to continue for at least the next few years, making it likely that AI capabilities will continue to expand rapidly. The AI systems that already exist today can be expected to have larger-scale effects on societies as they are employed in a multitude of ways. Future systems can be expected to be even more impactful and to transform societies and economies around the world. While there is enormous upside potential for these innovations, these systems also introduce risk.

Unfortunately, the striking capabilities of modern AI systems also enable new potential harms to people and society, both accidental and intentional. Current AI models can reproduce harmful biases in their training data,\footnote{Ondrej Bohdal et al., ``Fairness in AI and Its Long-Term Implications on Society'' (arXiv, 2023), \href{https://doi.org/10.48550/arXiv.2304.09826}{arXiv:2304.09826}.} evoke privacy concerns,\footnote{Karl Manheim and Lyric Kaplan, ``Artificial Intelligence: Risks to Privacy and Democracy,'' \emph{Yale Journal of Law and Technology} 21 (2019): 106–88, \url{https://yjolt.org/artificial-intelligence-risks-privacy-and-democracy}.} lack transparency,\footnote{Zihao Li, ``Why the European AI Act Transparency Obligation Is Insufficient,'' \emph{Nature Machine Intelligence} 5, no. 6 (2023): 559–60, \url{https://doi.org/10.1038/s42256-023-00672-y}.} and introduce new vulnerabilities in critical systems.\footnote{Phil Laplante and Ben Amaba, ``Artificial Intelligence in Critical Infrastructure Systems,'' \emph{Computer} 54, no. 10 (2021): 14–24, \url{https://doi.org/10.1109/MC.2021.3055892}.} AI systems may be particularly susceptible to \emph{misuse}—when AI is used for unethical ends such as the creation of disinformation, cyber-attacks, and scams.\footnote{Pranshu Verma, ``They Thought Loved Ones Were Calling for Help. It Was an AI Scam.,'' \emph{Washington Post}, March 5, 2023, \url{https://www.washingtonpost.com/technology/2023/03/05/ai-voice-scam/}; Josh A. Goldstein et al., ``Generative Language Models and Automated Influence Operations: Emerging Threats and Potential Mitigations'' (arXiv, 2023), \href{https://doi.org/10.48550/arXiv.2301.04246}{arXiv:2301.04246}. See also Miles Brundage et al., ``The Malicious Use of Artificial Intelligence: Forecasting, Prevention, and Mitigation'' (arXiv, 2018), \href{https://doi.org/10.48550/arXiv.1802.07228}{arXiv:1802.07228}. Unfortunately, misuse cannot be reliably distinguished from other uses at a technical level. For example, it might be perfectly legitimate to use an AI to find cyber vulnerabilities within a ``white hat'' cybersecurity firm, but that same process undertaken by a different actor could be deemed misuse.} These dangers are heightened by the possibility that AI systems sometimes acquire ``emergent capabilities,'' which surprise even their creators.\footnote{Jason Wei et al., ``Emergent Abilities of Large Language Models'' (arXiv, 2022), \href{https://doi.org/10.48550/arXiv.2206.07682}{arXiv:2206.07682}. Other work argues that emergent capabilities are primarily a measurement issue, though it remains true that models can surprise their creators, especially when capabilities are built on top of models after release. See Rylan Schaeffer, Brando Miranda, and Sanmi Koyejo, ``Are Emergent Abilities of Large Language Models a Mirage?'' (arXiv, 2023), \href{https://doi.org/10.48550/arXiv.2304.15004}{arXiv:2304.15004}. Unexpectedly dangerous capabilities arose following small changes to an existing system for molecule design, as detailed in Fabio Urbina et al., ``Dual Use of Artificial-Intelligence-Powered Drug Discovery,'' \emph{Nature Machine Intelligence} 4, no. 3 (2022): 189–91, \url{https://doi.org/10.1038/s42256-022-00465-9}.} Leading scientists and technologists have argued that if ever-more-powerful AI systems continue to be built, those systems could in time be capable of causing extraordinary damage to human society and may even threaten the extinction of humanity.\footnote{Center for AI Safety, ``Statement on AI Risk.''} The high potential for these systems, which are essentially digital files, to be copied, stolen, or misused against the will of their creators reinforces the intuition that even \emph{creating} them could be hazardous.\footnote{Richard Ngo, Lawrence Chan, and Sören Mindermann, ``The Alignment Problem from a Deep Learning Perspective'' (arXiv, 2023), \href{https://doi.org/10.48550/arXiv.2209.00626}{arXiv:2209.00626}; Center for AI Safety, ``Statement on AI Risk.''; Yoshua Bengio, ``How Rogue AIs May Arise,'' May 22, 2023, \url{https://yoshuabengio.org/2023/05/22/how-rogue-ais-may-arise/}.} An array of prominent experts, including top-tier AI researchers like Yi Zeng,\footnote{\emph{Artificial Intelligence: Opportunities and Risks for International Peace and Security - Security Council, 9381st Meeting} (United Nations Security Council, 2023), \url{https://media.un.org/en/asset/k1j/k1ji81po8p}.} Stuart Russell,\footnote{Stuart Russell, \emph{Human Compatible: Artificial Intelligence and the Problem of Control} (Penguin, 2019).} Geoffrey Hinton,\footnote{Cade Metz, ``‘The Godfather of A.I.’ Leaves Google and Warns of Danger Ahead,'' \emph{The New York Times}, May 1, 2023, \url{https://www.nytimes.com/2023/05/01/technology/ai-google-chatbot-engineer-quits-hinton.html}.}
and Yoshua Bengio\footnote{Yoshua Bengio, ``Slowing Down Development of AI Systems Passing the Turing Test,'' April 5, 2023, \url{https://yoshuabengio.org/2023/04/05/slowing-down-development-of-ai-systems-passing-the-turing-test/}; Bengio, ``How Rogue AIs May Arise.''}
as well as tech industry leaders
like Sam Altman,\footnote{Sam Altman, ``Machine Intelligence, Part 1,'' February 25, 2015, \url{https://blog.samaltman.com/machine-intelligence-part-1}.} Elon Musk,\footnote{Samuel Gibbs, ``Elon Musk: Artificial Intelligence Is Our Biggest Existential Threat,'' \emph{The Guardian}, October 27, 2014, \url{https://www.theguardian.com/technology/2014/oct/27/elon-musk-artificial-intelligence-ai-biggest-existential-threat}.} and Bill Gates,\footnote{Kevin Rawlinson, ``Microsoft’s Bill Gates Insists AI Is a Threat,'' \emph{BBC News}, January 29, 2015, \url{https://www.bbc.co.uk/news/31047780}.} have highlighted the need for society to take these risks seriously. Calls for regulation have also come from civil society\footnote{AI Now Institute, ``2023 Landscape Executive Summary,'' 2023, \url{https://ainowinstitute.org/general/2023-landscape-executive-summary}; Ardi Janjeva et al., ``Strengthening Resilience to AI Risk: A Guide for UK Policymakers'' (Centre for Emerging Technology and Security, 2023), \url{https://www.longtermresilience.org/post/paper-launch-strengthening-resilience-to-ai-risk-a-guide-for-uk-policymakers}.} and key firms, including Microsoft,\footnote{Microsoft, ``Governing AI: A Blueprint for the Future,'' 2023, \url{https://query.prod.cms.rt.microsoft.com/cms/api/am/binary/RW14Gtw}.} Google,\footnote{Kent Walker, ``A Policy Agenda for Responsible AI Progress: Opportunity, Responsibility, Security'' (Google, May 19, 2023), \url{https://blog.google/technology/ai/a-policy-agenda-for-responsible-ai-progress-opportunity-responsibility-security/}.} and OpenAI.\footnote{Altman, Brockman, and Sutskever, ``Governance of Superintelligence.''}

Addressing these problems requires governance, not just technical innovation. It may be crucial, for instance, that institutions restrain competition among firms and states in order to avoid dangerous ``race to the bottom'' interactions. Unrestrained competition among firms will pressure them to minimize their investments in safety.\footnote{Amanda Askell, Miles Brundage, and Gillian Hadfield, ``The Role of Cooperation in Responsible AI Development'' (arXiv, 2019), \href{https://doi.org/10.48550/arXiv.1907.04534}{arXiv:1907.04534}.} Similarly, if states think they can give their firms a competitive edge through lax regulation, a similar race to the bottom on regulatory standards can develop among states. 

At the domestic level, regulatory discussions are well underway in a number of states, including China, the EU, the US, and the UK. It remains to be seen how similar these regulatory approaches will turn out to be, but some differences are already apparent. The EU has been more concerned to protect privacy rights than the US has, for instance, as was already apparent in the EU’s General Data Protection Regulation (GDPR).\footnote{Anu Bradford, \emph{The Brussels Effect: How the European Union Rules the World} (Oxford University Press, 2020), ch.~2.} In spite of these differences, however, it is likely that some minimal best practice standards will emerge, particularly for addressing shared risks to public safety. 

Domestic AI regulations can ensure that these best practices are implemented, including potential licensing for AI firms and data centers,\footnote{The capability of cutting-edge AI models tends to scale smoothly with the amount of compute used in their training. Placing a strict cap on the amount of compute that a model can be trained with can thus help to manage the risk that unprecedentedly large models pose to society. Due to the huge cost of frontier models, such a cap, if carefully designed, would only affect the leading AI firms, not startups or other companies working in the space.} liability for AI firms, chain-of-custody accounting for aspects of the compute supply chain, model evaluations,\footnote{``OpenAI Evals'' (OpenAI, 2023), \url{https://github.com/openai/evals}.} and appropriate third party auditing of AI models.\footnote{Jakob Mökander et al., ``Auditing Large Language Models: A Three-Layered Approach,'' AI and Ethics, 2023, \url{https://doi.org/10.1007/s43681-023-00289-2}.} Regulation may need to apply to both model development and model deployment since it may be impossible to fully prevent unauthorized use of and access to a model once it has been developed.\footnote{Anderljung et al., ``Frontier AI Regulation''; Sabrina Küspert, Nicolas Moës, and Connor Dunlop, ``The Value Chain of General-Purpose AI'' (Ada Lovelace Institute, 2023), \url{https://www.adalovelaceinstitute.org/blog/value-chain-general-purpose-ai/}.} Containing an already trained model poses greater challenges due to its potential for wide proliferation. Furthermore, the compute requirements—or ``compute moat''—are considerably higher for model development than for deployment. This barrier is most pronounced during the training phase, underscoring the effectiveness of governance measures being in place before and throughout this stage of AI development. Domestic regulation can strongly incentivize firms within the jurisdiction to ensure that their AI systems perform according to societal expectations. Domestic governance will not be sufficient on its own, however.

\section{The Need for International Governance of Civilian AI}

International governance of AI is needed because AI poses international risks and successfully governing AI will require regulatory action by many states. States of different development levels and AI companies have different but overlapping interests in the development of international AI governance.

    \begin{wrapfigure}{R}{0.6\textwidth}
    \vspace{-7.8mm}
    \centering
    \tcbincludegraphics[arc=0mm,colback=white,left=0mm,right=0mm,width=\linewidth,graphics options={trim=3mm 2mm 1mm 19mm, clip}]{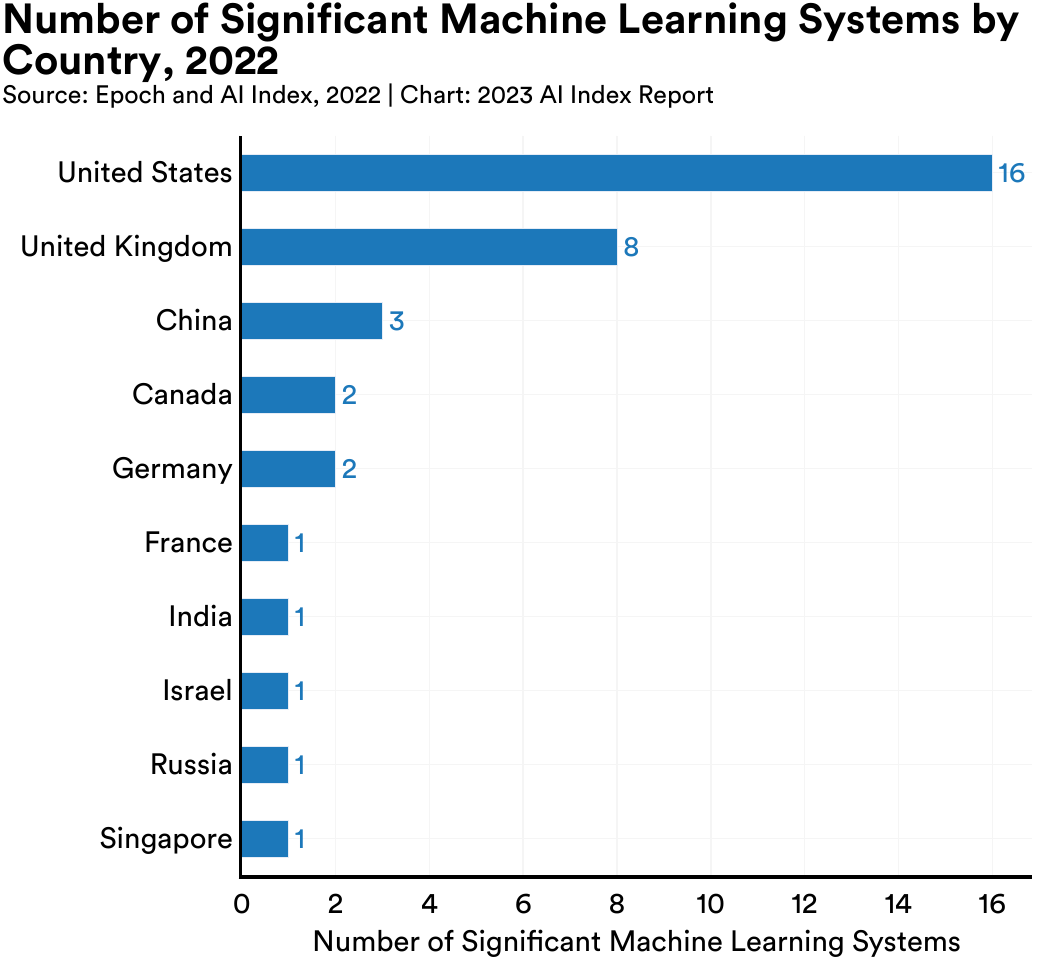}
    
    \parbox{\linewidth}{\setstretch{0.9}\begin{footnotesize}\emph{Source}: AI Index Report, 2023. Data from Epoch and AI Index, 2022. Criteria for including a model in the data set includes factors such as state-of-the-art improvements, historical significance, and the number of citations.\protect\footnotemark\ All models counted here were released in 2022.\end{footnotesize}\\[-5pt]}
    \caption{\textbf{Number of Significant Machine Learning Systems by Country, 2022. \\}}
    \label{figure1}
    \vspace{-1cm}
    \end{wrapfigure}
    \footnotetext{Epoch, “Parameter, Compute and Data Trends in Machine Learning,” 2022, \url{https://epochai.org/data/pcd}; Nestor Maslej et al., “Artificial Intelligence Index Report 2023” (Institute for Human Centered AI, 2023), \url{https://aiindex.stanford.edu/wp-content/uploads/2023/04/HAI_AI-Index-Report_2023.pdf}.}

Many AI risks can cross political borders. Internet-based digital services span the globe, making it possible for AI hazards, including accidents and misuse, to immediately harm people around the world.\footnote{Cybercrime is an important problem in this sphere. For example, see Julian Hazell, ``Large Language Models Can Be Used To Effectively Scale Spear Phishing Campaigns'' (arXiv, 2023), \href{https://doi.org/10.48550/arXiv.2305.06972}{arXiv:2305.06972}.} Theft is also a concern, since AI models are essentially digital files that can be copied exactly—and they can be easily moved via the internet or consumer-grade storage devices. Furthermore, AI models can contribute to the proliferation of dangerous weapon systems, including biological or chemical weapons, that have effects across borders.\footnote{Urbina et al., ``Dual Use of Artificial-Intelligence-Powered Drug Discovery.''} Future AI systems might even pose existential risks to humanity, thus making AI safety a central concern for all states.\footnote{Center for AI Safety, ``Statement on AI Risk.''}

\begin{figure}
    \centering
    \tcbincludegraphics[arc=0mm,left=1mm,right=1mm,colback=white,width=\linewidth,graphics options={trim=2mm 1mm 38mm 16mm, clip}]{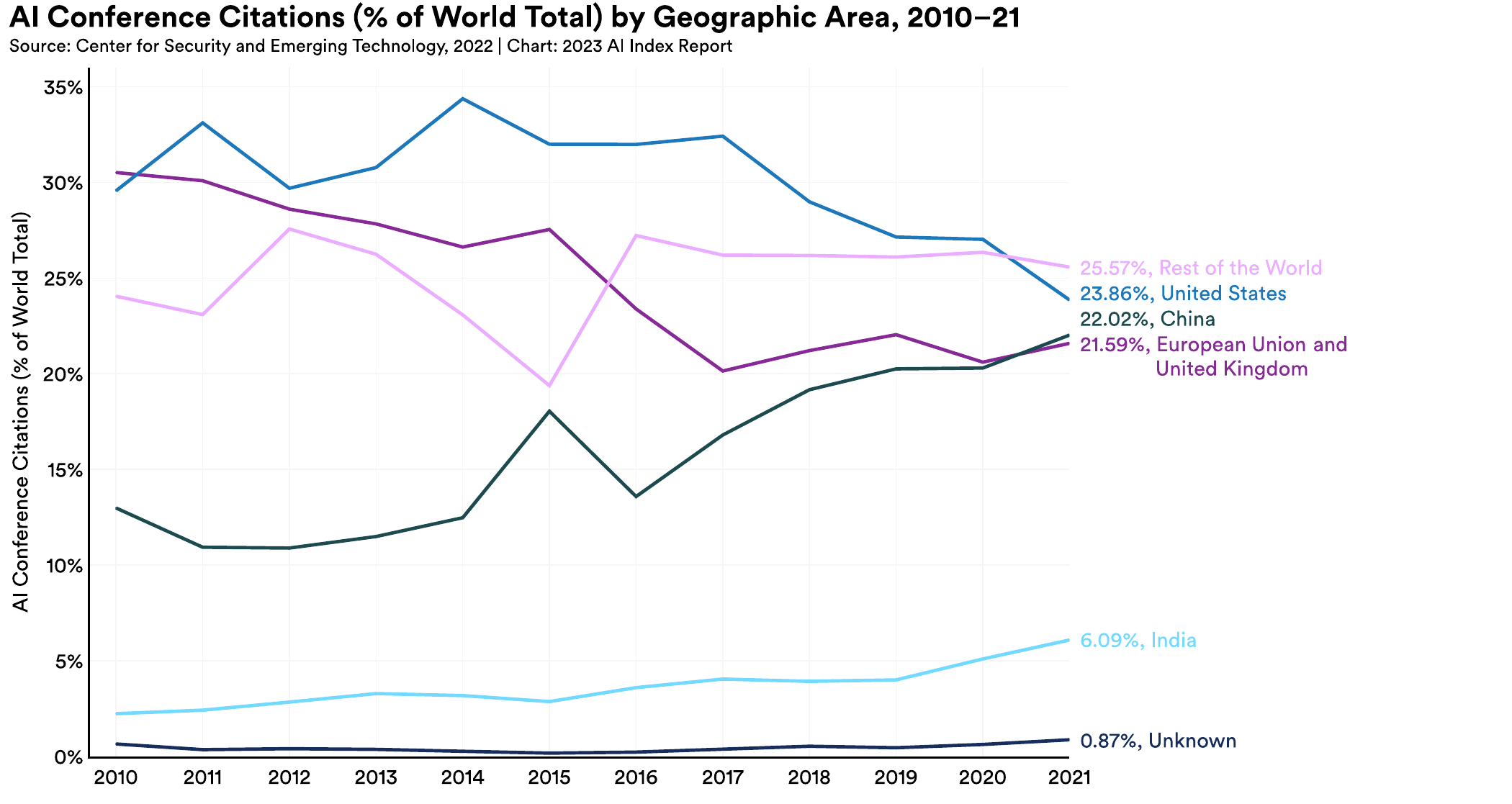}
    \parbox{\linewidth}{\vspace*{1pt}\begin{footnotesize}\emph{Source}: AI Index Report, 2023. Data from Center for Security and Emerging Technology, 2022.\end{footnotesize}\\[-1pt]}
    \caption{\textbf{AI Conference Citations (\% of World Total).}}
    \label{figure2}    
\end{figure}

The AI industry is also highly international, and governance will require action from many states. States at the forefront of the AI revolution include the US, the UK, and China (see Figure~\ref{figure1})—and many others have the potential to advance rapidly (see also Figure~\ref{figure2}). The compute supply chain—one of the key inputs for advanced AI—is also highly international.\footnote{It is unclear to what extent future AI advances will require continued access to the newest generation of chips, since algorithmic advances may allow teams with prior-generation compute hardware to build AI systems with the potential to do harm. In later sections, we discuss ``scaling laws'' which relate to this question.}

If left unaddressed, international competitive pressures can drive significant inefficiencies and dangers. Attempts to protect national AI industries might include trade barriers or lax regulations. Such moves would fragment AI regulation and make it more difficult to trade AI-related products and services across borders, thus making the AI industry and regulatory  system significantly less efficient. Even more concerning, such approaches could eventually lead states to weaken regulations in order to provide advantages for their firms. Absent strong regulations, corporations would primarily respond to market pressures and thus cut corners on safety—a situation that could produce some of the worst dangers of AI. 

Key actors have different but overlapping reasons to want harmonized AI regulations. Leading states would prefer that AI be safe and that they retain access to global markets. Similarly, other states would want AI to be safe and would furthermore desire improved access to AI technology.\footnote{Both of these goals would be served by strong, harmonized AI regulations. Later, we also discuss how regulatory harmonization is compatible with other approaches for sharing the benefits of AI.} All states face the difficulty of regulating an extremely complex technology, making it advantageous for them to pool some of their expertise and regulatory access.\footnote{Global supply chains can make AI products nearly inscrutable for regulatory agencies unless the agencies can rely on each other’s standards.} Corporations have multiple reasons to support regulatory harmony. A fragmented regulatory landscape results in higher costs from tailoring products to each jurisdiction. A mismatch of regulatory strength can cause firms in jurisdictions with strong regulation to lobby for harmonizing regulations across jurisdictions, since they would prefer a ``level playing field.''\footnote{Such an incentive is believed to have contributed to the support that US industry provided for the Montreal Protocol. See Elizabeth P. Barratt-Brown, ``Building a Monitoring and Compliance Regime Under the Montreal Protocol,'' \emph{Yale Journal of International Law} 16 (1991): 519–70, \url{https://openyls.law.yale.edu/handle/20.500.13051/6255}.} Furthermore, AI firms may also support strong international regulation in order to avoid transnational contagion effects such as a regulatory backlash caused by a high-profile failure of AI technology. A single major failure of AI anywhere in the world could both frighten investors and cause publics to associate the technology with dangerous outcomes. An important example of this effect can be seen in the history of civilian nuclear energy, where nuclear disasters such as Three Mile Island, Chernobyl, and Fukushima had a significant effect on global perceptions and adoption of the technology.\footnote{Bulat Aytbaev et al., ``Don’t Let Nuclear Accidents Scare You Away from Nuclear Power,'' Bulletin of the Atomic Scientists, August 31, 2020, \url{https://thebulletin.org/2020/08/dont-let-nuclear-accidents-scare-you-away-from-nuclear-power/}.}

These benefits of an international regulatory approach do \emph{not} imply that states must agree on ``one-size-fits-all'' regulation. Societal values differ, and these values should be expressed in national policies. We have already mentioned the differing approaches of the EU, US, and China towards privacy regulation, for instance. In spite of these differences, however, states can build consensus on the need to address a set of shared risks and adopt a set of minimal best practices for doing so. Surveying the risk landscape, threats to public safety are one area where all states have a great deal of shared values and interests. Within that area, many of the most significant risks are associated with the large, general systems at the cutting edge of capabilities that we term frontier AI. We therefore focus on this set of risks, but we also note that the approach to international civilian regulation that we describe later in the paper can be applied more broadly when international actors reach consensus on broader sets of issues.

In sum, the particular characteristics of AI as a technology and as an industry require that its governance be highly international. Unless states are able to achieve regulatory harmony, economic and regulatory competition could exacerbate risks across the international community. Harmonized international regulations for AI in targeted areas, such as with respect to threats to public safety, would reduce risks while also benefiting states, AI corporations, and people around the world.

\subsection{Reasons to Focus on Civilian AI}

AI will eventually be employed by every segment of human society, including militaries. Since militaries control the majority of humanity’s most dangerous capabilities, military AI will inevitably be an extraordinarily important domain for humanity to govern. Moreover, the particular mix of AI and militaries may be particularly concerning.\footnote{Forrest E. Morgan et al., ``Military Applications of Artificial Intelligence: Ethical Concerns in an Uncertain World'' (RAND Corporation, 2020), \url{https://www.rand.org/pubs/research\_reports/RR3139-1.html}.} Competition among state militaries can become much more severe than market competition among firms. In such a competitive environment, states may take significant risks in their pursuit of new capabilities.\footnote{For example, Manhattan Project scientists believed that there was a small risk that the Trinity test—the first nuclear explosion—would ignite the Earth’s atmosphere, killing all life on Earth. See Toby Ord, ``Lessons from the Development of the Atomic Bomb'' (Centre for the Governance of AI, 2022), \url{https://www.governance.ai/research-paper/lessons-atomic-bomb-ord}.} In particular, states that are not at the technological forefront may have the strongest incentives to cut corners on safety.\footnote{Eoghan Stafford, Robert Trager, and Allan Dafoe, ``Safety Not Guaranteed: International Strategic Dynamics of Risky Technology Races,'' Working Paper (Centre for the Governance of AI, 2022), \url{https://www.governance.ai/research-paper/safety-not-guaranteed-international-strategic-dynamics-of-risky-technology-races}.} 

\begin{figure}[ht]
    \centering
    \tcbincludegraphics[arc=0mm,left=0mm,right=0mm,colback=white,width=0.98\linewidth,graphics options={trim=0pt 0pt 30mm 13mm, clip}]{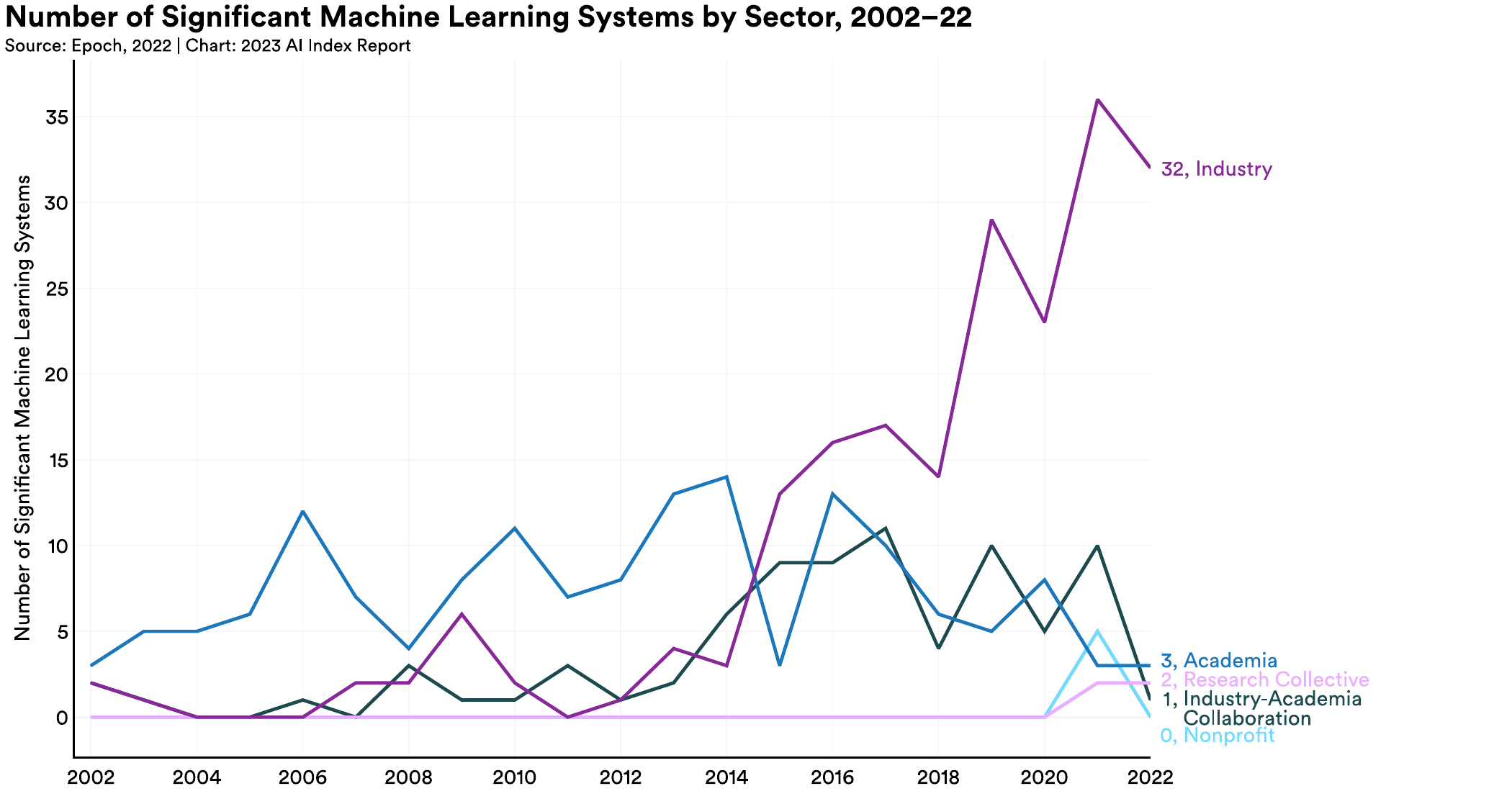}
    \parbox{\linewidth}{\footnotesize\vspace{0pt}\emph{Source}: 2023 AI Index Report. Data from Epoch 2022. Using the same data as Figure~\ref{figure1}, this plot shows how industry has come to a position of dominance in cutting-edge machine learning systems.\protect\footnotemark }
    \vspace{2pt}
    \caption{\textbf{Significant Machine Learning Systems by Sector.}}
    \label{figure3}    
    \end{figure}

In spite of these dangers, we contend that regulating civilian AI should be the first priority for three principal reasons.\footnote{Regulating state military use of AI is also an urgent problem, and it should be pursued in parallel.} First, unlike with technologies like nuclear weapons and the internet, which originated in state programs, the leading edge of AI development appears to be dominated by the private sector—at least for now. In 2022, the overwhelming majority of cutting-edge AI models were produced by private industry, continuing a decade-long trend of increasing industry dominance (see Figure~\ref{figure3}). These trends have been undergirded by private investments in AI far beyond what states appear to have invested so far.\footnote{For example, in 2022, the United States government is estimated to have invested about $\$$3 billion in AI, while US firms invested about $\$$47 billion. Maslej et al., ``Artificial Intelligence Index Report 2023,'' p 189, 286–88.} In sum, most cutting-edge AI research today is likely being done within firms, thus making civilian AI the primary domain in which advanced AI systems are being developed and released.

Second, regulating some aspects of international civilian AI appears feasible \emph{today}. The capabilities of recent AI products have triggered serious discussion of AI governance by civil society,\footnote{Future of Life Institute, ``Pause Giant AI Experiments: An Open Letter,'' March 22, 2023, \url{https://futureoflife.org/open-letter/pause-giant-ai-experiments/}; Center for AI Safety, ``Statement on AI Risk.''} industry,\footnote{Microsoft, ``Governing AI: A Blueprint for the Future''; Walker, ``A Policy Agenda for Responsible AI Progress''; Altman, Brockman, and Sutskever, ``Governance of Superintelligence.''} and politicians.\footnote{The White House, ``Readout of White House Meeting with CEOs on Advancing Responsible Artificial Intelligence Innovation''; António Guterres, ``Secretary-General Urges Broad Engagement from All Stakeholders towards United Nations Code of Conduct for Information Integrity on Digital Platforms,'' United Nations, June 12, 2023, \url{https://press.un.org/en/2023/sgsm21832.doc.htm}.} Domestic AI regulation is being actively discussed in all leading AI states;\footnote{``Oversight of A.I.''; Seaton Huang et al., trans., ``Translation: Measures for the Management of Generative Artificial Intelligence Services (Draft for Comment) – April 2023,'' April 12, 2023, \url{https://digichina.stanford.edu/work/translation-measures-for-the-management-of-generative-artificial-intelligence-services-draft-for-comment-april-2023/}; UK Department for Science, Innovation and Technology, ``AI Regulation: A Pro-Innovation Approach,'' March 2023, \url{https://www.gov.uk/government/publications/ai-regulation-a-pro-innovation-approach}; ``Proposal for a Regulation of the European Parliament and of the Council Laying down Harmonised Rules on Artificial Intelligence (Artificial Intelligence Act) and Amending Certain Union Legislative Acts'' (2021), \url{https://eur-lex.europa.eu/legal-content/EN/TXT/?uri=celex\%3A52021PC0206}.} extending these processes with an international component is a natural—and necessary—next step in these discussions. Indeed, international AI regulation has already come to the fore,\footnote{For example, international governance was mentioned multiple times during the US Senate hearing ``Oversight of A.I.: Rules for Artificial Intelligence'' and was part of the joint statement by Rishi Sunak and key AI CEOs in ``PM Meeting with Leading CEOs in AI: 24 May 2023,'' GOV.UK, May 24, 2023, \url{https://www.gov.uk/government/news/pm-meeting-with-leading-ceos-in-ai-24-may-2023}.} with a Global Summit on AI Safety being planned for late 2023 and the Secretary-General of the United Nations convening a High-Level Advisory Board on Artificial Intelligence to provide options for global governance of AI by the end of this year.\footnote{``UK to Host First Global Summit on Artificial Intelligence''; Secretary-General's remarks to the Security Council on Artificial Intelligence, 18 July 2023.}

    \footnotetext{Epoch, ``Parameter, Compute and Data Trends in Machine Learning''; Maslej et al., ``Artificial Intelligence Index Report 2023.''}

    \begin{figure}[ht]
    \begin{centering}
    \tcbincludegraphics[arc=0mm,left=0mm,right=0mm,bottom=0mm,colback=white,width=\linewidth,graphics options={trim=1mm 2mm 9mm 13mm, clip}]{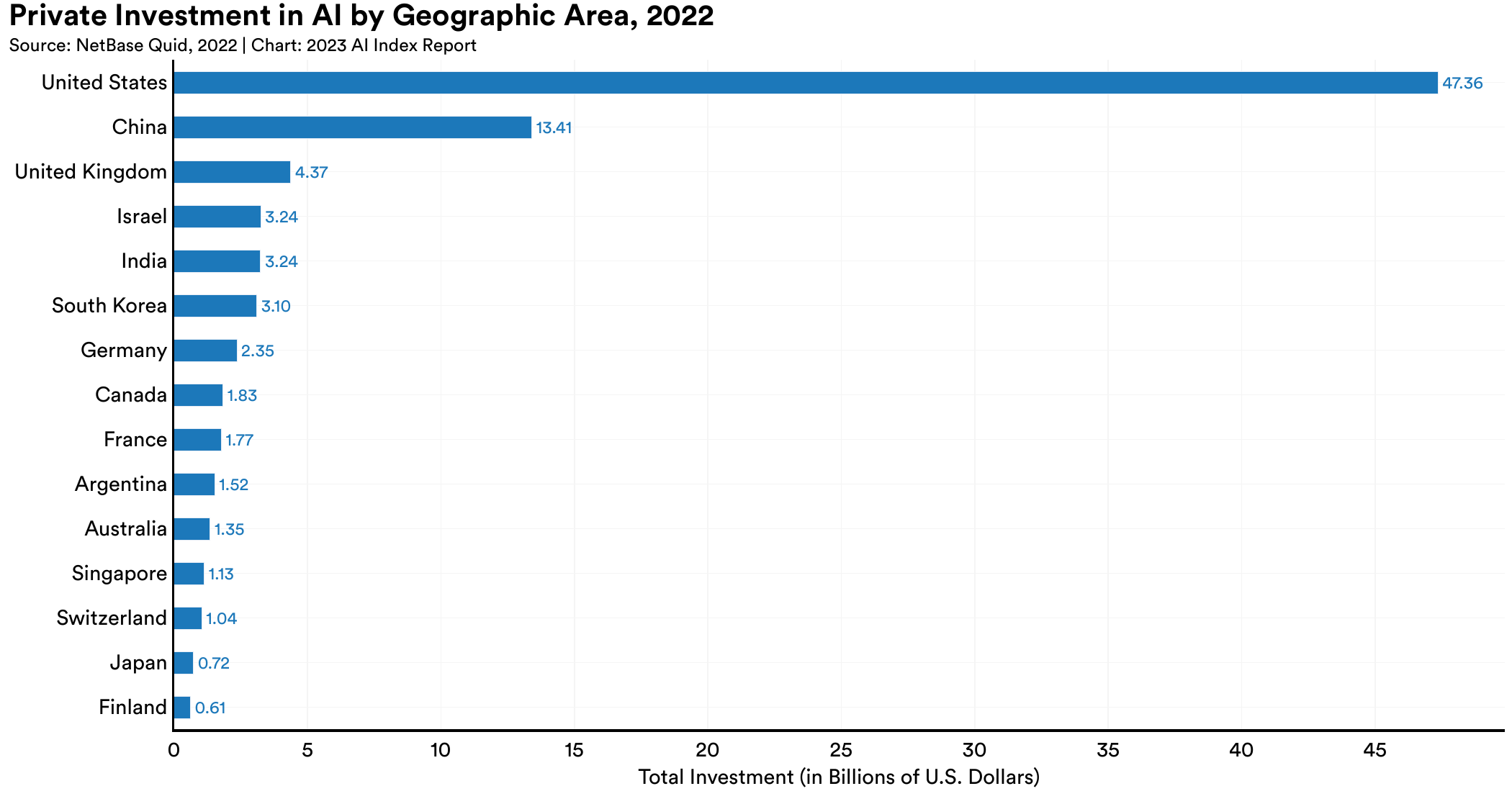}
   \end{centering} 
   \begin{flushleft}\begin{footnotesize}\emph{Source}: 2023 AI Index Report. Data from NetBase Quid, 2022.\end{footnotesize}\end{flushleft}
    \vspace{4.5pt}
   \centering \caption{\textbf{Private Investment in AI by Geographic Area, 2022.}}
    \label{figure4}
    \end{figure}

Third, arms control is often unsuccessful and tends to be difficult and slow. While the international governance of military uses of AI (such as an ``IAEA for AI,'' which would apply to both civilian and military uses)\footnote{Altman, Brockman, and Sutskever, ``Governance of Superintelligence.''} is a desirable goal which should be pursued, achieving it will likely require prolonged effort.\footnote{John Newhouse, \emph{War and Peace in the Nuclear Age} (New York: Alfred A. Knopf, 1988); Thomas M. Nichols, \emph{No Use: Nuclear Weapons and U.S. National Security} (Philadelphia: University of Pennsylvania Press, 2013).} Thus, the challenges of arms control should not hold up regulation of civilian AI.

Regulating civilian AI is also likely to be an important first step toward eventually regulating state and military uses of AI. Civilian technologies and supply chains typically, and perhaps increasingly, undergird most military technology, thus allowing civilian regulation to have an indirect influence on the safety and reliability of military AI. At the very least, civilian safety and reliability regulations can become a \emph{de facto} standard to which military AI will be held. Furthermore, practical experience with specific governance processes developed for civilian AI—including technical monitoring techniques—can inform the development of analogous processes for military AI.

\section{Civilian AI Governance: Components and Trade-offs}

Setting up an international civilian governance ecosystem for frontier AI involves a series of institutional design choices that must be tailored to the technology. The features of the technology that must be taken into account include the extent to which it is ``dual-use,'' the societal consequences if the rules are broken, the opportunities for control of inputs into the technology, and the nature of opportunities for discovering violations of the regime. We have already discussed some of the factors that must be considered in the case of AI, but we summarize some of the most important ones here for reference:

\begin{itemize}
    
    \item The current capabilities of the technology are reliably associated with scale in both compute and data.\footnote{Joel Hestness et al., ``Deep Learning Scaling Is Predictable, Empirically'' (arXiv, 2017), \href{https://doi.org/10.48550/arXiv.1712.00409}{arXiv:1712.00409}; Jared Kaplan et al., ``Scaling Laws for Neural Language Models'' (arXiv, 2020), \href{https://doi.org/10.48550/arXiv.2001.08361}{arXiv:2001.08361}; Tom Henighan et al., ``Scaling Laws for Autoregressive Generative Modeling'' (arXiv, 2020), \href{https://doi.org/10.48550/arXiv.2010.14701}{arXiv:2010.14701}; Pablo Villalobos, ``Scaling Laws Literature Review'' (Epoch, 2023), \url{https://epochai.org/blog/scaling-laws-literature-review}.} In recent years, model size has been doubling every ten months.\footnote{Sevilla et al., ``Compute Trends Across Three Eras of Machine Learning.''} Algorithmic advancements can imply drastic reductions in the amount of compute required to train models with a given level of capabilities.\footnote{Erdil and Besiroglu, ``Algorithmic Progress in Computer Vision.''.}

    \item The capabilities of frontier AI are potentially unpredictable.\footnote{Wei et al., ``Emergent Abilities of Large Language Models.''; Deep Ganguli et al., ``Predictability and Surprise in Large Generative Models,'' in \emph{Proceedings of the 2022 ACM Conference on Fairness, Accountability, and Transparency}, FAccT ’22 (ACM, 2022), 1747–64, \url{https://doi.org/10.1145/3531146.3533229}. Other work has shown that part of the unpredictability of model capabilities may be a function of the measures one examines. See Schaeffer, Miranda, and Koyejo, ``Are Emergent Abilities of Large Language Models a Mirage?''.}
    \item An emerging field of ``model evaluations'' is developing the capacity to test new AI systems, at all stages of development and deployment, for threats to public safety and other harms.\footnote{Ethan Perez et al., ``Discovering Language Model Behaviors with Model-Written Evaluations'' (arXiv, December 19, 2022), \href{https://doi.org/10.48550/arXiv.2212.09251}{arXiv:2212.09251}; ``OpenAI Evals.''; ARC Evals, ``Update on ARC’s Recent Eval Efforts,'' March 17, 2023, \url{https://evals.alignment.org/blog/2023-03-18-update-on-recent-evals/}; OpenAI, ``Our Approach to AI Safety,'' April 5, 2023, \url{https://openai.com/blog/our-approach-to-ai-safety}; Toby Shevlane et al., ``Model Evaluation for Extreme Risks'' (arXiv, 2023), \href{https://doi.org/10.48550/arXiv.2305.15324}{arXiv:2305.15324}.}

    \item Some types of model evaluations will require a range of types of model access, from input/output access to knowledge of model internals (e.g. gradients, embeddings, and other internal parameters) and training environments.\footnote{Ben Bucknall, Toby Shevlane, and Robert Trager, ``Structured Access for Third-Party Safety Research on Frontier AI Models Investigating Researchers’ Model Access Requirements'' (Working Paper, n.d.); Inioluwa Deborah Raji et al., ``Closing the AI Accountability Gap: Defining an End-to-End Framework for Internal Algorithmic Auditing'' (arXiv, 2020), \href{https://doi.org/10.48550/arXiv.2001.00973}{arXiv:2001.00973}; Toby Shevlane, ``Structured Access: An Emerging Paradigm for Safe AI Deployment'' (arXiv, 2022), \href{https://doi.org/10.48550/arXiv.2201.05159}{arXiv:2201.05159}; Shevlane et al., ``Model Evaluation for Extreme Risks.''}

    \item The technical expertise needed to develop standards exists largely in a small number of private-sector organizations. Any new body will need to draw on this expertise in a structured way in order to develop standards.\footnote{The ICAO's \href{https://www.icao.int/about-icao/AirNavigationCommission/Pages/anc-technical-panels.aspx\#:$\sim$:text=1\%29\%20Develop\%20and\%20update\%20strategies\%20and,to\%20facilitate\%20worldwide\%20coordination\%20of\%20implementation\%3B\&text=1\%29\%20Develop\%20and\%20update,worldwide\%20coordination\%20of\%20implementation\%3B\&text=and\%20update\%20strategies\%20and,to\%20facilitate\%20worldwide\%20coordination}{Air Navigation Commission} and its technical panels are an example of drawing on industry expertise.} 

    \item Advanced forms of AI are a safety-critical technology. Violations of the regime have the potential to cause large-scale societal harms.\footnote{Ngo, Chan, and Mindermann, ``The Alignment Problem from a Deep Learning Perspective''; Center for AI Safety, ``Statement on AI Risk.''; Bengio, ``How Rogue AIs May Arise.''}

    \item Harmful forms of proliferation resulting from the access required for the model evaluations being developed is a concern.

    \item There are three essential inputs into the technology: algorithms, data, and compute. Of these, compute may be the easiest to control internationally, in part because it is ``rivalrous''—possession by one actor excludes possession by another.

    \item The compute and data requirements for frontier models are such that a relatively small number of private actors have the capability to create them—at least so far. However, systems built on top of existing large models can transform the capabilities of those models and thus require regulation.

    \item Determined state actors, and potentially others, are capable of exfiltrating AI systems.

    \item The computing resources required to run advanced models are much less than those required to create them.\footnote{Ying Sheng et al., ``FlexGen: High-Throughput Generative Inference of Large Language Models with a Single GPU'' (arXiv, 2023), \href{https://doi.org/10.48550/arXiv.2303.06865}{arXiv:2303.06865}.} A much larger number of actors are capable of running models, if they gain access to them, than are capable of creating novel systems.

\end{itemize}    

In developing an international civilian AI governance ecosystem to account for these features of the technology, institutional design choices can be grouped into four overlapping areas: the standard setting ecosystem, monitoring, incentives for compliance, and governance of the institutions themselves.\footnote{Ho et al., ``International Institutions for Advanced AI.''} We highlight key trade-offs in each before describing in detail one potential international civilian governance ecosystem for advanced AI.

\subsection{The Standard Setting Ecosystem}

International standard setting ecosystems exist across many industries, such as accounting, finance, forestry, aviation, and electronics. These ecosystems have one or more international standard setting authorities that usually interact with local government standards bodies. In the case of the International Organization for Standardization (ISO), which has published more than 24,500 standards across many industries, national standards bodies make up the voting membership of the organization. Many industries have their own standard setting bodies that often work together with the ISO and contribute to its standards.

One common ecosystem model includes a number of ``certification bodies''—firms or government entities that audit industry firms and projects, certifying them on the basis of standards developed by the industry standards body. These certification bodies may then themselves be audited by an ``accreditation body'' that provides oversight of the certification bodies. For example, more than forty organizations worldwide are certification bodies for Forest Stewardship Council (FSC) standards.\footnote{``FSC-Accredited Certification Bodies,'' Forest Stewardship Council UK, accessed June 22, 2023, \url{https://uk.fsc.org/fsc-accredited-certification-bodies}.} These FSC certification bodies are accredited by Assurance Services International, the sole accreditation body for the FSC standards.\footnote{``Certification System,'' Forest Stewardship Council, accessed June 22, 2023, \url{https://connect.fsc.org/certification/certification-system}.} An international certification system for farm feed additives operates similarly, though it has several accreditation bodies.\footnote{``Accreditation Bodies,'' FAMI QS, accessed June 22, 2023, \url{https://fami-qs.org/certified-organisations/accreditation-bodies/}.} 

Another model involves a single organization that both develops standards and performs the auditing or monitoring function. This is the approach taken, at least in part, in the maritime, aviation, and nuclear industries.\footnote{The institutions managing these industries are the International Maritime Organization (IMO), International Civil Aviation Organization (ICAO), and International Atomic Energy Agency (IAEA) respectively.} In many cases, multiple certifications, such as from state governments as well as internationally recognized certification bodies, are required for a project to move forward. 

In some cases, different regions have separate standard setting ecosystems. For example, the European Union has its own regional institutions for standard setting, including the European Committee for Standardization. Alternatively, a state may perform an auditing function on its own to supplement auditing by an international body. The US Federal Aviation Administration has such a program to supplement International Civil Aviation Organization (ICAO) certification.\footnote{Federal Aviation Administration, ``International Aviation Safety Assessment (IASA) Program,'' accessed June 22, 2023, \url{https://www.faa.gov/sites/faa.gov/files/about/initiatives/iasa/FAA\_Initiatives\_IASA.pdf}.} Many of the ICAO’s other 193 member countries have domestic civil aviation standards bodies of varying levels of capability, and aviation safety is governed and implemented through a much broader ecosystem of government, non-governmental, and private-sector actors. The ICAO sets a framework of minimum standards globally, conducts audits, and offers capacity-building support. But much of the direct safety impact on the aviation industry is not caused directly by the ICAO; it emerges from an ecosystem of national (e.g. the Federal Aviation Administration (FAA) in the US, the Civil Aviation Authority (CAA) in the UK, etc.), regional (e.g. European Civil Aviation Conference), and international (e.g. the International Air Transport Association (IATA)) organizations. These work in concert with private-sector airlines and airports, creating mutual reinforcement of safety standards that is greater than the sum of its parts.\footnote{Interview with a senior British aviation safety official.}

\begin{mybox} \textbf{The International Civil Aviation Organization (ICAO)} is a UN agency that audits state aviation oversight systems and publishes each state’s level of compliance with ICAO standards in a report. The United States Federal Aviation Administration (FAA), alongside other national regulatory bodies, gives force to ICAO standards.\footnotemark\ If the FAA determines that a country’s oversight system is not in compliance with ICAO standards, it can prohibit that country’s airlines from operating in the US. Other jurisdictions, such as China and the European Union, have adopted related policies. Thus, any country whose airlines seek to operate in some of the world’s largest aviation markets must meet at least some of the ICAO safety oversight standards.
\end{mybox}
\footnotetext{For example, the US Transportation Security Administration (TSA) is obligated to inspect foreign airports that send flights to the US. United States Government Accountability Office, ``Aviation Security: TSA Strengthened Foreign Airport Assessments and Air Carrier Inspections, but Could Improve Analysis to Better Address Deficiencies,'' 2017, \url{https://www.gao.gov/assets/gao-18-178.pdf}. See also Federal Aviation Administration, ``International Aviation Safety Assessment (IASA) Program.''}

Another question is whether the standard setting body is based in a new or an existing institution. A healthy international industry governance ecosystem involves a web of checks and balances. 

\subsection{The Type of Monitoring}

Monitoring is different depending upon the needs of each industry. In some cases, chain-of-custody\footnote{Chain-of-custody auditing attempts to ``trace, verify, document and aggregate the history, location and application of every item in the whole supply chain,'' and is particularly common in industries that attempt to demonstrate that goods brought to market were sustainably sourced. For a discussion of different approaches, see ``Supply Chain Model: Chain of Custody,'' Deloitte, n.d., \url{https://www2.deloitte.com/nl/nl/pages/sustainability/articles/chain-of-custody.html}.} auditing is required, for instance to demonstrate that wood brought to market has been responsibly sourced.\footnote{``Chain of Custody Certification,'' Forest Stewardship Council, accessed July 13, 2023, \url{https://fsc.org/en/chain-of-custody-certification}.} The FSC system employs chain-of-custody certification to ensure that wood marketed with the FSC certification is sustainably harvested and then traded only among certified institutions. A chain-of-custody approach is particularly useful when the monitored product is \emph{fungible}—making it difficult or impossible to distinguish between certified and uncertified products. Digital assets can have this property, but preventive measures are available.\footnote{Techniques exist for marking the ownership of digital assets and for guaranteeing that they have not been tampered with. New standards are also emerging which allow for a robust chain of custody accounting for digital resources. For example, see ``Overview,'' Coalition for Content Provenance and Authenticity, accessed July 14, 2023, \url{https://c2pa.org/}.} The International Electrotechnical Commission (IEC), by contrast, manages peer reviews among its conformity assessment bodies (to ensure that standards are upheld) and stipulates that jurisdictions cannot undertake duplicate testing (to prevent jurisdictions from making the process more onerous or politicized).\footnote{``How the Global IEC Conformity Assessment Systems Operate: A Network of Trust,'' International Electrotechnical Commission, accessed June 22, 2023, \url{https://www.iec.ch/conformity-assessment/how-global-iec-ca-systems-operate}.} 

\begin{mybox} \textbf{The International Maritime Organization (IMO)} is a UN agency focused on the global shipping industry. Among its many functions is an audit scheme whereby signatory states are audited for their compliance with IMO standards. While the IMO itself has no enforcement powers of its own, the recommendations it generates from its audits can be highly motivating for states. If a state falls out of compliance with key IMO standards, the economic consequences can be ``serious and far reaching,'' as their ships can be denied entry to—or detained in—ports in other jurisdictions, and ships from signatory states face the prospect of costly inspections and delays if they interact with the ports of non-compliant states.\footnotemark\ Thus, the IMO serves as a crucial central clearinghouse for compliance information, which is then used to inform the domestic enforcement processes within signatory states.
\end{mybox}
\footnotetext{``Frequently Asked Questions on Maritime Security,'' International Maritime Organization, accessed June 9, 2023, \url{https://www.imo.org/en/OurWork/Security/Pages/FAQ.aspx}.}

A key distinction is whether international monitoring targets jurisdictions or firms. In the forestry example, certification bodies audit firms and their projects. In the aviation and maritime examples, the ICAO or the IMO audits jurisdictions to ensure that regulations are consistent with international standards.\footnote{``Evolving ICAO’s Universal Safety Oversight Audit Programme: The Continuous Monitoring Approach,'' \emph{ICAO Journal} 65, no. 4 (2010): 24–25, \url{https://www.icao.int/safety/CMAForum/Shared\%20Documents/6504\_en-1.pdf}; International Maritime Organization, ``Framework and Procedures for the IMO Member State Audit Scheme,'' December 5, 2013, \url{https://wwwcdn.imo.org/localresources/en/OurWork/MSAS/Documents/MSAS/Basic\%20documents/A.1067(28)\%20Framework\%20and\%20Procedures.pdf}.} The FATF performs similar audits and also evaluates whether state authorities effectively carry out the regulations on the books.\footnote{``Mutual Evaluations,'' Financial Action Task Force, accessed July 13, 2023, \url{https://www.fatf-gafi.org/en/topics/mutual-evaluations.html}.}

\subsection{The Incentives for Compliance}

Many standards are enforced through markets that demand certification as information about product quality. For example, FSC-certified wood can fetch a higher price because customers can be more confident that it has been sustainably harvested.\footnote{Standards adopted locally can also have international effects, for instance through the so-called ``Brussel’s Effect.'' The de facto and de jure effects of the General Data Protection Regulation (GDPR) on companies’ privacy policies and government regulation in at least 120 countries are the clearest examples of the Brussels effect. In order to maintain access to the European market, digital companies like Apple, Facebook, Google and Microsoft adopted EU policy on privacy when the GDPR was adopted in 2016. See Bradford, \emph{The Brussels Effect}; and Charlotte Siegmann and Markus Anderljung, ``The Brussels Effect and Artificial Intelligence: How EU Regulation Will Impact the Global AI Market'' (arXiv, 2022), \href{https://doi.org/10.48550/arXiv.2208.12645}{arXiv:2208.12645}.}

In other cases, there is direct cross-border enforcement by states. When the IAEA detects a violation of a nuclear safeguard, for instance, it can make a referral to the United Nations (UN) Security Council, sometimes resulting in military action by member states. Such actions are of course contentious, involving divergent interests of world powers, and such processes take time.

Other important incentives for compliance occur through ties to the trade regime. A state may mandate that international certification is required for the import of a technology. Similarly, some countries require jurisdictions to be in compliance with ICAO standards in order for planes originating from those jurisdictions to enter their airspace.\footnote{The ICAO provides oversight of member states in five regulatory areas (including aviation legislation and operating instructions) and three areas of implementation (including licensing and resolution of safety issues). The ICAO also has the authority to issue ``mandatory information requests'' about defined aspects of a state’s safety oversight system. Audits may identify a ``Significant Safety Concern''—a possible deficiency in ``the ability of the audited State to properly oversee its airlines (air operators); airports; aircraft; and/or air navigation services provider under its jurisdiction.'' See ``Frequently Asked Questions about USOAP,'' International Civil Aviation Organization, n.d., \url{https://www.icao.int/safety/CMAForum/Pages/FAQ.aspx}.} In addition, many countries use ICAO standards as a baseline and have additional safety requirements for planes originating from another jurisdiction to enter their airspace. So, a decision to stop flying is a bilateral one but is embedded in an international framework. For example, in 2015, the United Kingdom stopped all flights to and from Egypt's Sharm el Sheikh International Airport following a Metrojet charter flight to St Petersburg which crashed in the Sinai desert shortly after take-off from the airport. This creates strong and layered incentives. Other countries may look to the bilateral decision and make their own decision to stop flights. The ICAO may also engage in such scenarios not as an enforcer, but to investigate and make proposals for how the country falling short of standards can improve (and increase the confidence of the international community).\footnote{Interview with a senior UK aviation safety official.}

In the US, the Federal Aviation Administration (FAA)’s International Aviation Safety Assessment (IASA) program investigates whether jurisdictions are in compliance with ICAO standards. If a country is not in compliance, the FAA can prohibit that country’s airlines from operating in the US.\footnote{``International Aviation Safety Assessment (IASA) Program,'' Federal Aviation Administration, n.d., \url{https://www.faa.gov/about/initiatives/iasa}. See also, Morgan Simpson and Robert Trager, ``Cooperation in Safety-Critical Industries: Lessons for AI from Aviation and Nuclear,'' Working Paper, n.d.} For its part, China stipulates that flight crew licenses issued by other countries are only valid for operating within China if those licenses meet ICAO standards.\footnote{The Civil Aviation Administration of China (CAAC) has adopted Article 181 of its \href{http://www.caac.gov.cn/en/ZCFG/MHFL/201509/P020150901511659239730.pdf}{Civil Aviation Law of the People's Republic of China}, p.75, which states: ``The civil aircraft certificates of airworthiness and certificates of competency and licences of crew members issued [by a foreign state] shall be recognized as valid by the Government of the People's Republic of China, provided that the requirements under which such certificates or licences were issued or rendered valid shall be equal to or above the minimum standards established by the International Civil Aviation Organization.'' The European Union has a similar regulation; see ``Regulation (EU) 2018/1139 of the European Parliament and of the Council'' (European Union, July 4, 2018), \url{https://eur-lex.europa.eu/legal-content/EN/TXT/?uri=CELEX\%3A32018R1139}. Note that the ICAO does not itself issue licenses; it issues standards for licensing by state civil aviation authorities. See ``Personnel Licensing FAQ,'' International Civil Aviation Organization, n.d., \url{https://www.icao.int/safety/airnavigation/pages/peltrgfaq.aspx}.} 

On the export side, multilateral export control regimes deny export of particular technologies to jurisdictions that do not meet requirements, often geopolitical ones. According to a recent study on the effectiveness of international treaties, instruments that have a trade component are much likelier to produce their intended economic and social effects.\footnote{Steven J. Hoffman et al., ``International Treaties Have Mostly Failed to Produce Their Intended Effects,'' \emph{Proceedings of the National Academy of Sciences} 119, no. 32 (August 9, 2022): e2122854119, \url{https://doi.org/10.1073/pnas.2122854119}.} In Appendix A, we discuss the compatibility with international trade law of import and export controls on AI products and precursors.

\subsection{The Nature of Institutional Governance}

The governance of institutions that regulate the standards ecosystem—the composition of the governing board, for instance—itself involves an important set of trade-offs and will be determined by the actors who create the ecosystem. Thus, the desired form of governance of a proposed institution is an important factor in deciding who should be invited to participate, and at what stage, in discussions of the institution’s creation.

One key question is whether the standard setting and monitoring organization is an independent non-governmental organization, an independent intergovernmental organization, or part of another intergovernmental body, such as the United Nations or a regional organization. While some standards ecosystems are convened by intergovernmental processes, others originate from collaborations among private entities, including firms and representatives of affected groups. This often represents a trade-off between speed and effectiveness on the side of private entities and greater legitimacy on the side of broad intergovernmental oversight. Private entities like the Internet Corporation for Assigned Names and Numbers (ICANN) are often criticized for their lack of legitimate oversight, especially, but far from exclusively, by states who wish to have more influence over its decisions. On the other hand, ICANN has proven effective at administering certain aspects of the internet and keeping access open to all. In the case of independent organizations, it is often important to have a permanent secretariat that is independent of the organization’s membership. This appears to facilitate the organization’s credibility, legitimacy, and effectiveness.\footnote{See Ranjit Lall, \emph{Making International Institutions Work: The Politics of Performance} (Cambridge, UK: Cambridge University Press, 2023), \url{https://doi.org/10.1017/9781009216265}. Whether or not an AI standard setting and monitoring organization forms part of a larger existing organization, it will need to work with existing standard setting initiatives. These include the ISO’s ISO/IEC FDIS 42001 standards, which are currently in development, as well as governance processes at the Council of Europe, the OECD, and elsewhere.}

The composition of governing boards and assemblies is a particularly important design consideration. Mandating that a broad set of stakeholders take part can assuage legitimacy concerns, even for private entities and public-private partnerships.\footnote{For discussion of the range of governance options for standard setting institutions, see Kenneth W. Abbott and Duncan Snidal, ``The Governance Triangle: Regulatory Standards Institutions and the Shadow of the State,'' in \emph{The Politics of Global Regulation}, ed. Walter Mattli and Ngaire Woods (Princeton University Press, 2009), 44–88, \url{https://doi.org/10.1515/9781400830732.44}. On public-private partnerships, see also Oliver Westerwinter, ``Transnational Public-Private Governance Initiatives in World Politics: Introducing a New Dataset,'' \emph{The Review of International Organizations} 16, no. 1 (2021): 137–74, \url{https://doi.org/10.1007/s11558-019-09366-w}.} The Forest Stewardship Council (FSC), for instance, is a model in this regard. Its General Assembly, the highest decision-making body, is composed of members from three ``chambers'': environmental, social, and economic. The chambers are each composed of ``private enterprises, NGOs, international organisations, indigenous groups, and educational institutions'' and each chamber has equal voting power in the assembly.\footnote{``Governance,'' Forest Stewardship Council UK, n.d., \url{https://uk.fsc.org/governance}.} These structures attempt to mitigate industry capture of standards bodies, which is an ever-present concern. The IAEA provides yet another model. It guarantees board seats to the ten nations that are judged by the previous board to be most advanced in atomic energy technology.\footnote{International Atomic Energy Agency, ``The Statute of the IAEA'' (n.d.), \url{https://www.iaea.org/about/statute}, Article VI.} 

It can sometimes be important to enable particular stakeholders to exercise greater influence over decisions, even if there is some cost in terms of equity and legitimacy; without such influence, these actors may not participate in the regime at all. There are a variety of options for enabling certain states to exercise greater influence, including: (1) weighted voting (e.g. based on GDP), (2) permanent seats on the executive board, and (3) consensus decision-making (which in practice tends to give powerful states more influence).

The composition and structure of the governing bodies of the international institutions that regulate advanced AI will be particularly important. Powerful states—and powerful labs—have some divergent interests and will advocate for differing policies. States with less advanced AI industries may seek to use international institutions to discover technological secrets, as is believed to have occurred in the case of the IAEA.\footnote{Matthew Fuhrmann, \emph{Atomic Assistance: How ``Atoms for Peace'' Programs Cause Nuclear Insecurity} (Ithaca, NY: Cornell University Press, 2012); Christoph Bluth et al., ``Civilian Nuclear Cooperation and the Proliferation of Nuclear Weapons,'' \emph{International Security} 35, no. 1 (2010): 184–200, \url{https://www.jstor.org/stable/40784651}; Elisabeth Roehrlich, \emph{Inspectors for Peace: A History of the International Atomic Energy Agency} (Baltimore: JHU Press, 2022), p. 361.} Less advanced states will also be wary of exclusion from processes that govern technologies with global effects. If governance of the institutions is too contentious, the ecosystem will be sclerotic and not achieve its objectives.

\section{A Jurisdictional Certification Approach to International Civilian AI Governance}

An international civilian AI governance regime has three essential elements: standards, monitoring, and enforcement. We describe an approach that provides for each and is closely related to approaches used in other industries. It is perhaps most closely related to the civil aviation, maritime, and financial activities regimes centered around the ICAO, IMO, and FATF.

We presume that domestic regulators in advanced AI states have taken up the challenge of beginning to regulate AI development and deployment. Given the range of conversations that have already begun on these topics, it appears likely that domestic regimes combining licensing (or a close substitute) and liability will emerge in the coming years.\footnote{Proposal for a Regulation of the European Parliament and of the Council Laying down Harmonised Rules on Artificial Intelligence (Artificial Intelligence Act) and Amending Certain Union Legislative Acts; ``Oversight of A.I.''; Helen Toner et al., ``How Will China’s Generative AI Regulations Shape the Future? A DigiChina Forum,'' DigiChina, April 19, 2023, \url{https://digichina.stanford.edu/work/how-will-chinas-generative-ai-regulations-shape-the-future-a-digichina-forum/}.} 

In such an environment, a first step to one form of internationalization would be for the leading AI regulators to collaborate with other countries to set up mirroring regulatory agencies or capacities,\footnote{There is a debate over whether AI regulation should be centralized in a single agency or whether competencies should exist across governments, often focused on particular application domains. We do not address these issues here, although we think it likely, for practical reasons, that the centralized approach will be more effective. One reason is simply the burden on legislatures if they attempted to create differentiated standards across use cases instead of delegating some of this authority to agencies.} building on existing initiatives such as the US-EU Trade and Technology Council, the Global Partnership on AI, and the G7/OECD processes. The regulators could then coordinate in harmonizing licensing and liability regimes, setting up an international standard setting and monitoring organization, and ensuring international incentives for compliance with international standards. 

\subsection{International AI Organization (IAIO): Standards Harmonization and Jurisdictional Certification}

Even as advanced AI states set up AI regulatory capacities, they should share information on best practices with other states and encourage them to implement their own analogous regulatory capabilities. Leading regulatory organizations, working with relevant government agencies, could determine what technical information can be shared with the nascent regulatory bodies of foreign states. Furthermore, these agencies could facilitate the exchange of technical experts. Throughout this process, states and civil society must forge consensus about minimum standards for appropriate civilian development and deployment of AI. 

A next step is the creation of an international standard setting and jurisdictional monitoring organization. This would facilitate standards harmonization and compliance, just as similar organizations do in other industries.\footnote{This approach is particularly similar to the governance of international aviation (ICAO) and shipping (IMO).} Without such a body, even if all states developed regulatory agencies, it would remain unclear whether regulations are in harmony and whether states are successfully regulating AI within their jurisdictions. 

A group of aligned states could invest in the creation of an International Artificial Intelligence Organization (IAIO). The IAIO would partner with national regulators in developing an international set of standards for data centers, AI firms, and regulatory jurisdictions. It would certify \emph{jurisdictions}—as opposed to firms—for standards compliance, including the jurisdictions’ statutory adoption of the international regulatory standards and their capacity to reliably enforce their regulations.\footnote{To loosely extend the civil aviation analogy, we might analogize these activities to the TSA’s foreign airport assessments and air carrier inspections for compliance with ICAO standards. United States Government Accountability Office, ``Aviation Security: TSA Strengthened Foreign Airport Assessments and Air Carrier Inspections, but Could Improve Analysis to Better Address Deficiencies.'' Note that such an arrangement does not require the granting of specific authorities through treaty. The incentive to conform to IAIO standards derives from national interests and the connection of IAIO standards to trading standards.} This would likely include an assessment of whether a country’s regulatory system achieves a defined set of outcomes, similar to the FATF’s effectiveness assessment of 11 ``immediate outcomes'' related to money laundering and terrorist financing.\footnote{Financial Action Task Force, ``Methodology for Assessing Compliance with the FATF Recommendations and the Effectiveness of AML/CFT Systems,'' 2023, \url{https://www.fatf-gafi.org/en/publications/Mutualevaluations/Fatf-methodology.html}. Note that FATF’s version of jurisdictional ``certification'' is to place jurisdictions with ``weak measures to combat money laundering and terrorist financing'' on either a ``black'' or ``grey'' list. See ``‘Black and Grey’ Lists,'' Financial Action Task Force, n.d., \url{https://www.fatf-gafi.org/en/countries/black-and-grey-lists.html}.} We discuss a specific version of data center operator and AI-firm regulation, based on licensing, below.

\subsection{Enforcement via Conditional Market Access}

\subsubsection*{Imports}

Enforcement of the international regime would start with market access that is made conditional on certification. Similar to what is done in other industries, states can adopt safety regulations indicating that they will only allow the import or sale of relevant AI products whose supply chains involve only IAIO-certified jurisdictions.\footnote{See Appendix A for discussion of AI product and precursor trade restrictions’ compliance with international trade law.} This would provide a strong incentive for jurisdictions around the world to implement and enforce these standards. Given the challenges of controlling software, when compared to physical products, this will likely require the continued evolution of export control frameworks so that they can be effective in this context. If most large markets adopted such rules, IAIO compliance would become extremely desirable for any state developing commercial AI technologies. See the sidebars above on the ICAO and IMO ecosystems for examples of this in the aviation and naval industries.

\subsubsection*{Exports}

Exports can be similarly shaped by IAIO certification. Participating states could add IAIO certification as a requirement for export of AI inputs, models, and products, ensuring that non-compliant jurisdictions cannot easily gain access to advanced capabilities or inputs into AI production processes. Participating states would thus refrain from exporting sensitive technology to non-certified states.\footnote{Certification would likely be a necessary, but not a sufficient, condition for export of AI supply chain technologies as the current AI supply chain export restrictions appear to be determined by additional factors.} 

One form this has taken in other sectors such as advanced missile capabilities and nuclear, biological, chemical, and conventional weapons is a multilateral export control regime.\footnote{These export control regimes are the Missile Technology Control Regime, the Nuclear Suppliers Group, the Australia Group, and the Wassenaar Arrangement.} Such regimes help member states keep sensitive materials and technologies out of the hands of dangerous actors and geopolitical rivals. One particularly interesting aspect of these regimes is that they sometimes have to manage the proliferation of technologies and materials that are quite generic or general purpose, such as particular chemicals,\footnote{``Export Control List: Chemical Weapons Precursors,'' The Australia Group, accessed June 9, 2023, \url{https://www.dfat.gov.au/publications/minisite/theaustraliagroupnet/site/en/precursors.html}.} chemistry equipment,\footnote{``Dual-Use Chemical Manufacturing Facilities and Equipment,'' The Australia Group, accessed June 9, 2023, \url{https://www.dfat.gov.au/publications/minisite/theaustraliagroupnet/site/en/dual\_chemicals.html}.} and biology equipment.\footnote{``Control List of Dual-Use Biological Equipment and Related Technology and Software,'' The Australia Group, accessed June 9, 2023, \url{https://www.dfat.gov.au/publications/minisite/theaustraliagroupnet/site/en/dual\_biological.html}.} If a similar multilateral export control regime were to be developed around AI, some best practices from prior regimes can be used. Such regimes facilitate the exchange of information about which exports are potentially sensitive and how export control decisions are being made. They also ensure that non-members cannot easily ''shop around'' for a willing exporter (known as the ''no undercut'' policy). A multilateral export control regime could become an evolving mechanism by which certified states clarify their shared understanding of how they will limit the spread of potentially harmful AI capabilities.\footnote{Evidence suggests that international agreements that contain enforcement provisions linked to trade and finance laws are the most likely to achieve their objectives. See Hoffman et al., ``International Treaties Have Mostly Failed to Produce Their Intended Effects.''}

\subsection{Requiring Enforcement for Certification}

One technique for increasing the strength of enforcement is to require states to implement the trade restrictions described above in domestic law as a condition for certification. This approach to enforcement strengthens incentives for states to join the agreement and to stay if they have already joined. Enforcement of international agreements typically requires one or more states to muster the political will to punish states that deviate from the agreement—thus often creating a free-rider problem which leads to weak enforcement. By contrast, requiring enforcement as a condition for certification turns that logic on its head to some extent. \emph{Enforcement} becomes the default outcome unless political capital is expended to modify international regulations. In a sense, \emph{avoiding enforcement} rather than enforcement is associated with a collective-action problem.

An agreement of this form is more robust, but that robustness may come at a cost. Launching this strengthened agreement among a group of core states would likely require greater political will than the weaker alternative—which merely requests that states embed enforcement provisions in their trade laws.\footnote{Although a countervailing factor is that political decision makers can legitimately claim to their constituencies that this agreement is a serious attempt to solve the problem of civilian AI governance.} Furthermore, while this stronger agreement has a greater ability to be self-enforcing, if a key state chose to exit the agreement, that action would have a chance of setting off a cascade of interactions wherein this stronger agreement would be downgraded to its weaker version—if it survived at all. A small number of key markets might trigger such a cascade upon their departure. Nevertheless, requiring enforcement provisions in domestic law has worked well in other domains, such as some aspects of maritime regulation overseen by the IMO.

\subsection{IAIO Jurisdictional Standards}

The international regulatory ecosystem that we sketch here is compatible with many approaches to jurisdictional standards and could have the benefit of allowing national governments flexibility on the precise regulatory mechanism with which they implement the standards. This approach would both preserve national sovereignty and likely be a pragmatic and flexible approach to developing a coherent global regulatory framework. 

Below, we sketch one approach to regulating frontier AI systems that begins with three forms of licensing, ensuring that regulators have oversight of all frontier systems being developed and that such systems are deployed in compliance with safety standards. 

The IAIO could create standards or specifications that require countries to implement a jurisdictional licensing regime for:

\begin{enumerate}
    \item \emph{Development and Deployment of Frontier Models} employing more than a certain amount of floating point operations, or FLOP (e.g. $> 10^{24}$ FLOP).\footnote{The amount of operations requiring regulatory scrutiny would need to be adjusted over time to account for algorithmic efficiency and other factors.} Firms would be required to submit information to domestic regulators in advance of system creation, including: ``model cards'' specifying training procedures and the data used, the computing hardware to be employed, and other aspects of the project.\footnote{Model cards summarize key information about the model. Margaret Mitchell et al., ``Model Cards for Model Reporting,'' in \emph{Proceedings of the Conference on Fairness, Accountability, and Transparency}, FAT$^\ast$’19 (ACM, 2019), 220–29, \url{https://doi.org/10.1145/3287560.3287596}.} The proposed systems would be evaluated by the firm or by third parties to understand the profile of risks it could pose to society. Regulators would have the technical ability to check that the proposed project was in fact run on the proposed hardware.\footnote{Yonadav Shavit, ``What Does It Take to Catch a Chinchilla? Verifying Rules on Large-Scale Neural Network Training Via Compute Monitoring'' (arXiv, 2023), \href{https://doi.org/10.48550/arXiv.2303.11341}{arXiv:2303.11341}.} Regulators would regulate access to the deployed model, including mandating security measures to prevent impermissible forms of fine tuning and structured querying, exporting, and unauthorized copying of model weights and code. 

    \item \emph{AI Firms} training models using more than a certain amount of compute. Licensing would be contingent upon demonstrating system security, following guidance on model sharing, documenting past compliance with development and deployment regulations, and other factors.

    \item \emph{Data Centers and Data Center Operators} above a certain capacity (e.g. $> 1,000$ data-center-quality chips).\footnote{It may be more effective to define data centers in terms of overall FLOP/s, without reference to specific chips. Regulation may also take into account the geographic concentration of chips (and interconnectivity bandwidth) in guarding against attempts to circumvent regulation by separating computing clusters in such a way that they could be used to train models in tandem but still fall below the size threshold for auditing. The requirement to track and report frontier model training would assist regulators in determining when actors may be splitting training across providers to avoid regulatory oversight.} These actors would be prohibited from providing access to computing power to unlicensed AI firms or for unlicensed projects. They would be required to (1) provide accountings of all data-center-quality chips purchased from fabricators, (2) have robust cybersecurity measures to protect frontier models from malicious attacks and adversarial actors, and (3) track and report when customers are training frontier models or accessing them for high-risk uses. Violating these requirements would result in penalties and potential loss of license.

\end{enumerate}

Alongside licensing, domestic legislators should create synergistic forms of liability to deter potential harms. In the US, for instance, existing tort law (enforced via lawsuits) and consumer protection law (enforced mainly via Federal Trade Commission (FTC) action) have significant limitations—even establishing legal standing to address many potential harms may be difficult. New statutes or rules should impose penalties for violating responsible development and release standards. 

Furthermore, the IAIO can also serve as a central node for information pertaining to AI governance which is not the purview of a specific state. For example, one function of the IAIO could be to track the location and ownership of key inputs to civilian AI such as AI-optimized computing hardware—akin to how the International Atomic Energy Agency (IAEA) facilitates tracking nuclear material. Empowered with access to information about such inputs, the IAIO would be much better positioned to make judgements about whether jurisdictions are fully abiding by the international standard. As we note above, such information may also be important to future efforts to regulate non-civilian AI. The organization could also collect and share information about emerging AI risks and assist regulators to develop and implement their regulatory regimes, as the ICAO and IMO do. 

\subsection{Governance of the IAIO}

Governance of the IAIO is a difficult issue because powerful countries, including among the five permanent members of the UN Security Council (P5), have somewhat divergent interests as regards regulation. These diverging interests could lead to conflicts over policy that would interfere with the organization's mission. For instance, some countries might see the monitoring role of the IAIO as an opportunity to gain insight into the capabilities of the most advanced firms in rival states. Yet, advanced firms and their home states would likely wish to minimize such unsanctioned information transfers.

These divergent interests contrast with the harmonized interests of the P5 in many other industries, including atomic power and aviation. The common interest of nuclear weapons states in preventing other states from acquiring nuclear weapons is clear in the context of the International Atomic Energy Agency (IAEA), for instance. These common interests of powerful countries ease the governance problem for both the IAEA and the ICAO.\footnote{Permanent members of the Security Council have convergent interests as well. Among these are preventing global catastrophes and ensuring that access to the technology does not enable smaller actors to threaten them, for instance through access to open-sourced models.}

Given the differing interests among powerful states, including the P5, it is unclear whether the IAIO should be a UN organization like the ICAO. Another option is for it to be an independent, non-profit organization, and a third option is a public-private partnership (PPP). The Internet Corporation for Assigned Names and Numbers (ICANN) is a model for independent, non-profit organization (see sidebar). A private organization or PPP would need to take significant steps to achieve broad legitimacy around the world. The initial board of the organization could reflect a balance of substantial representation from around the world and knowledge of technical issues. The Global Alliance for Vaccines and Immunization (Gavi), and the Global Fund to Fight AIDS, Tuberculosis and Malaria are examples of PPPs that have functioned effectively. These organizations include in their governing bodies governments as well as firms, NGOs, unions, and other non-state actors. 

\begin{mybox}
\textbf{The Internet Corporation for Assigned Names and Numbers (ICANN)} is another model of international standard setting and regulation. Unlike the ICAO, which is part of the United Nations, ICANN is a private, non-profit organization that regulates part of the technical backbone of the internet, including its domain name system. The International Organization for Standardization (ISO), which develops thousands of standards across many industries, is yet another model.\footnotemark\ Given the strained state of international relations, it is noteworthy that neither of these organizations were formed through treaties. Indeed, even the Organization for Security and Cooperation in Europe (OSCE), which monitors elections, was also formed without a treaty.
\end{mybox}
\footnotetext{Tim Büthe and Walter Mattli, \emph{The New Global Rulers: The Privatization of Regulation in the World Economy} (Princeton, NJ: Princeton University Press, 2011).}

Whatever the legal basis for establishing the IAIO, its governing board membership could replicate principles found in other international organizations. The board of the IAEA, for instance, has places reserved for the most advanced nuclear technology states—whomever they may be at the time. A similar approach for the IAIO might ameliorate the issue of divergent interests of powerful states by enabling the states with the most advanced AI industries to have greater say in governance of the organization. Over time, as additional states develop more advanced capabilities, the organization’s governance would adjust via a mechanism that evolves representation along with state capabilities. Such a mechanism would need to be balanced against maintaining a voice for non-frontier states affected by the technology.

\subsection{International Firm-Level Monitoring}

While it is reasonable to expect large industrial states such as the US and China to develop highly capable domestic regulatory agencies, such infrastructure would be infeasible for smaller or less developed states to create or maintain. To solve this problem, another element of the proposed IAIO would be the capability to monitor firms directly at the behest of the state which holds jurisdiction over those firms. In this role, the IAIO would fulfill part of the role of a domestic regulator by scrutinizing firms for compliance. Concrete enforcement (such as legal penalties) would be provided by the home state, but the IAIO could do the technical heavy lifting required to monitor and certify actors for compliance. At a country’s request, the organization would also provide assistance in building regulatory systems using international best practices, including through seconding experts and mobilizing assistance from international partners. This structure would help participating states achieve and maintain compliance with IAIO standards at a lower cost than if they created similar capabilities themselves. This could be particularly helpful for countries who would otherwise have difficulty participating in the IAIO due to resource constraints.\footnote{This approach bears some resemblance to how the IMO provides guidance and support for states that are trying to abide by IMO rules. The Vision Statement of the IMO audit framework is ``To promote the consistent and effective implementation of applicable IMO instruments and to assist Member States to improve their capabilities, whilst contributing to the enhancement of global and individual Member State's overall performance in compliance with the requirements of the instruments to which it is a Party.'' International Maritime Organization, ``Framework and Procedures for the IMO Member State Audit Scheme.''} Some countries may also choose to delegate monitoring to the IAIO to expedite their entry into the international market, especially if they lack the capacity to rapidly develop national regulatory capacities or if their national cybersecurity is not robust enough to safeguard sensitive data. As a central node in the network, the IAIO’s monitoring capabilities would also benefit from economies of scale, potentially resulting in cost benefits for all states monitored by the IAIO, along with more effective standardization.

This aspect of the IAIO would be a service to states wishing to cost-effectively demonstrate that they are fulfilling their oversight obligations.\footnote{As Hans Blix noted when he was Director General of the International Atomic Energy Agency, comparing Agency inspections to the regular inspections of an elevator company, ``If you had a sign saying that the owner of the house has inspected it, maybe there wouldn’t be the same credibility.'' See Roehrlich, \emph{Inspectors for Peace}, p. 380.} For a state to be in full compliance, the monitoring processes of the IAIO would need to be credibly connected to state enforcement mechanisms in order to ensure that firm compliance failures are addressed in a timely fashion.

International monitoring of firms via the IAIO would be expected to reduce costs in several ways. States using IAIO firm monitoring would avoid 1) the initial cost of building a domestic technical monitoring agency, 2) the ongoing costs of maintaining and updating that agency as IAIO standards evolve, and 3) cybersecurity costs as best practices evolve. Moreover, it might be less costly for the IAIO to certify jurisdictions that use IAIO firm monitoring. 

International firm monitoring via the IAIO also provides a way for states to gain additional certainty that other markets are being regulated fairly. IAIO firm monitoring would be set up to apply similar standards across states, thus creating a more level playing field. 

\begin{figure}[ht]
    \centering
    \includegraphics[width=\linewidth]{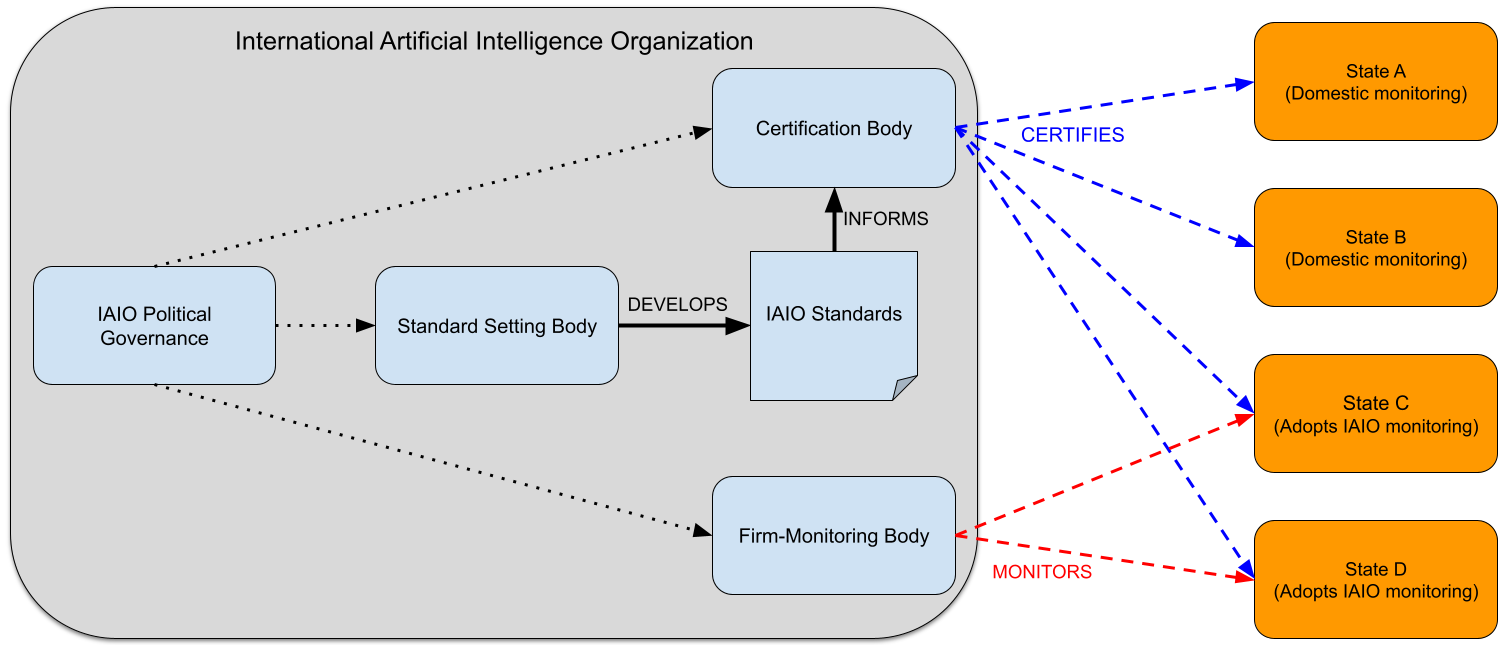}
    \vspace{3pt}
    \caption{\textbf{IAIO Authorities}}
    \label{figure5}
\end{figure}

The certification and firm-monitoring organs of the organization are represented in Figure~\ref{figure5}. The organization would also have the ability to assist jurisdictions in complying with IAIO standards. This tracks the ICAO and IMO examples, as these organizations provide substantial assistance to member states. Indeed, these organizations can mobilize international efforts to assist member states in resolving safety concerns.\footnote{One such case occurred following the ICAO audit of the Kyrgyz Republic in 2016. This effort included not only ICAO staff, but also technical experts from Georgia, Turkey, Ukraine, and the United States, who performed training on site, in addition to donated training courses in France, the UK, and Singapore. See ``Another ‘No Country Left Behind’ Success: A Significant Safety Concern Resolved!,'' International Civil Aviation Organization, n.d., \url{https://www.icao.int/EURNAT/Pages/news\_articles/NoCountryLeftBehind-success.aspx}.} The staff of these different organs of the IAIO could be elected to limited terms by the governing/executive body, with quotas to ensure broad regional representation.

In addition to domestic and IAIO monitoring, some states with frontier, proprietary capabilities may seek to monitor firm activities and regulations in other jurisdictions themselves. The aviation industry again provides an analogy. Alongside the ICAO, the US Federal Aviation Administration (FAA) operates its own International Aviation Safety Assessment of other states.\footnote{Note that this FAA program analyzes a country's ability, not the ability of individual air carriers, to adhere to international safety standards.} Such an approach could be desirable when a monitoring state has relevant technical insights about standards that it is unwilling to share with other states and IAIO personnel. Optionally, some states could be incentivized to accept firm-level monitoring, on the part of either the IAIO or a leading state, by making such oversight a condition for participation in the import and export control regimes. Such actions would constitute ``extraterritorial'' applications of laws and thus would likely be controversial; they might also undercut the IAIO regime by making IAIO certification less desirable in itself.

\begin{mybox}
\textbf{Elements of an international standards regime.}

\vspace{.5cm}

\begin{itemize}[itemsep=-11pt] 
    \item \emph{International AI Organization (IAIO)}, an independent, non-profit organization:
    \begin{itemize}
	\item Develops standards in cooperation with firms and national regulators
        \item Certifies regulatory \emph{jurisdictions} for standards compliance and enforcement capacity
        \item Optionally: Partners with states to monitor firms as a service\\
    \end{itemize} 
    \item \emph{Export Control Regime}:
    \begin{itemize}
        \item Optionally: IAIO certification is a necessary, not a sufficient, condition for receiving exports of advanced AI inputs
        \item Optionally: Ties export permissions to IAIO firm-level monitoring \\
    \end{itemize} 
    \item \emph{States}:
    \begin{itemize}
        \item Regulate domestic firms according to IAIO standards
	\item Monitor firms through a domestic agency or the IAIO
	\item Make information on domestic regulation available to the IAIO to achieve certification
	\item Maintain the export control regime
	\item Adopt import standards requiring any AI involved in product development to be trained in an IAIO-certified jurisdiction
	\item Encourage other states to support all aspects of the regime
	\item Develop national technical capacities for international AI-firm monitoring
	\item Optionally: In some cases, tie import standards to IAIO firm-level monitoring
	\item Optionally: Develop independent jurisdictional and firm-level certification programs on the model of the FAA.
    \end{itemize}
\end{itemize}
\end{mybox}

\subsection{Mitigating Proliferation Dangers from Governance Processes}

Governance regimes for powerful technologies must avoid furthering harmful forms of proliferation. The IAIO system described above may enable proliferation risks for at least two reasons. First, personnel within the IAIO system may learn technical secrets—such as algorithms, data engineering techniques, and key hyperparameters—either through fulfilling their official duties or through unofficial channels (including incidental occurrences or unauthorized action). Second, IAIO systems—as well as the domestic governance systems that they interact with—may collect and store data that could be copied (either on-site or via a cyber-attack).

The approach proposed here mitigates proliferation concerns by having local governments maintain responsibility for oversight of domestic firms. If an international body were scrutinizing the most advanced firms, proliferation through the monitoring process would be likely, given the forms of access to system development techniques that are likely to be required to ensure systems’ safety. We expect, however, that the states in which the most advanced firms are housed will manage their own regulatory processes in accordance with international standards.\footnote{It may also be possible to mitigate proliferation concerns through privacy/security-preserving techniques such as differential privacy. For an overview of promising research avenues, see Miles Brundage et al., ``Toward Trustworthy AI Development: Mechanisms for Supporting Verifiable Claims'' (arXiv, 2020), \href{https://doi.org/10.48550/arXiv.2004.07213}{arXiv:2004.07213}. For discussion of privacy-preserving monitoring in the nuclear safeguards verification regime, see Mauricio Baker, ``Nuclear Arms Control Verification and Lessons for AI Treaties'' (arXiv, 2023), \href{https://doi.org/10.48550/arXiv.2304.04123}{arXiv:2304.04123}.}

The chief proliferation concerns that this design raises, therefore, involve the information embodied in the technical standards themselves and in the technical knowledge required to develop standards. The IAIO’s standard setting process will need to be in close dialog with domestic regulators, industry players, and academics to ensure that international AI regulations are updated rapidly in accordance with advances in the field. Indeed, the rapid pace of development makes essential the rapid updating of standards and the rapid propagation of these updates down to domestic regulators. The IAIO could perform this standard setting function most effectively if it possessed full knowledge of the science behind model development, deployment, and standard setting. This would likely include knowledge of leading algorithms and conceptual approaches employed in the training of AI models.\footnote{Bucknall, Shevlane, and Trager, ``Structured Access for Third-Party Safety Research on Frontier AI Models Investigating Researchers’ Model Access Requirements.''} Yet, such knowledge on the part of the IAIO could make the organization a vector for harmful proliferation.

We believe this concern is significant but also that it should not be overstated. It is likely that standard setting processes at the IAIO will be able to evaluate and apply some technical model evaluation standards without furthering harmful proliferation. Testing and evaluation standards to prevent algorithmic bias, for example, likely fall largely into this category. Even in the case of model evaluations to prevent threats to public safety, some standards may not require or embody substantial knowledge of the technological frontier. For instance, it may not require substantial proprietary technical knowledge to set standards for evaluating whether AI systems can give ``instructions on how to carry out acts of terrorism,'' or coerce users in pursuit of objectives. The Alignment Research Center, which does have deep technical expertise, recently performed related evaluations on Anthropic and OpenAI systems.\footnote{See ARC Evals, ``Update on ARC’s Recent Eval Efforts'' and Shevlane et al., ``Model Evaluation for Extreme Risks.'' Standard setting information that could lead to harmful proliferation includes information about how models can be augmented with additional capabilities after they are released and specific ways AI systems could be used to cause harm.}

The existence of such standards implies that the IAIO can promote public welfare without furthering proliferation. Nevertheless, a challenge remains in that other useful standards may require frontier knowledge, particularly when it comes to evaluating the standards themselves. Consider, for instance, a standard of a level of FLOP above which models would require greater regulatory scrutiny and evaluation. Improvements in algorithms can change what can be achieved with a given amount of FLOP; thus, knowledge of the state of the art of algorithmic efficiency might be required to set such a standard effectively.\footnote{Note, however, that knowledge of algorithmic efficiency levels is different from knowledge of the algorithms themselves. Regulators might know the amount of compute used to train models as well as the benchmarks those models achieved, giving them the ability to estimate algorithmic efficiency—without knowledge of the algorithms.} 

There are a number of ways to address this issue. One is for the IAIO to adopt procedures similar to those the IAEA adopted for national intelligence after the disclosure of clandestine nuclear sites in Iraq following the Persian Gulf War. The IAEA decided to consider material shared by national intelligence services in evaluating states’ nuclear programs, but the organization also maintained its own ability to evaluate the veracity of shared material. The IAEA understood that states would be reluctant to share intelligence if doing so would endanger sources and methods. It therefore took steps to prevent this, including having staff sign nondisclosure agreements, restricting access to intelligence to small numbers of staff, punishing breaches of confidentiality, and implementing measures to increase cyber and physical information security. Moreover, ``informed states could provide their information via private briefings with the director general—whom powerful states ensure they trust during the selection process—and a select few staff members.''\footnote{Allison Carnegie and Austin Carson, ``The Disclosure Dilemma: Nuclear Intelligence and International Organizations,'' \emph{American Journal of Political Science} 63, no. 2 (April 2019): 269–85, \url{https://doi.org/10.1111/ajps.12426}. The authors argue that the measures taken by the IAEA to prevent the transfer of intelligence secrets to rivals led to greater sharing with the organization.} We believe that such measures will be helpful in some cases, but that states at the technological frontier, or states whose firms are, will probably be unwilling to share information in some cases where the public interest—absent harmful proliferation—would be served if they did share.\footnote{This challenge will be particularly acute if appropriate standards for civilian AI highlight technical capabilities that are dual-use. } 

In some cases, an option for the IAIO or similar organization would be to consider a standard recommended by a state without the technical explanation for why that standard is necessary. This would be similar to states revealing intelligence information to the IAEA without revealing sources and methods. As in that case, such information may be less effective at motivating change. 

As we have mentioned, another option for states that are unwilling to share frontier knowledge needed for effective standard setting and monitoring is to set up their own standardization and monitoring organizations separately from the IAIO. States might do this individually, as the United States does in the civil aviation sector with the FAA’s International Aviation Safety Assessment.

Overall, therefore, we should expect that some of the work of an international standards regime would not be subject to proliferation concerns. In these areas alone, the regime would likely improve public welfare. In other areas, actors will be more reticent to share information, and standard setting will be more contested.\footnote{Consider, as an example, an algorithmic advance that increased computing efficiency by 10x. Sharing this information with an international regulator would risk revealing it to competitors and adversarial governments. Note, however, that it might be less important to share the information with international regulators before the insight has begun to diffuse internationally. In the case of localized advances, domestic regulator awareness might be sufficient for effective oversight.} Procedural solutions similar to the IAEA’s handling of intelligence information can ameliorate these difficulties but likely will not fully solve them.

\subsection{Alternative Governance Approaches}

We have highlighted one approach to an international regime for civilian AI standard setting, monitoring, and enforcement, but other approaches to international governance should also be considered.\footnote{For a consideration of a range of options for international governance of frontier AI, see Ho et al., ``International Institutions for Advanced AI.''} We will briefly describe the key differences between each model and our proposed approach. See Table~\ref{table1final} for a summary.

\begin{table}[H]
    \renewcommand{\thetable}{1}
    \centering
    \tcbincludegraphics[arc=0mm,left=0mm, right=0mm, top=0mm, bottom=0mm, colback=white,width=\linewidth,graphics options={trim=8mm 8mm 8mm 15mm, clip}]{images/table1.png}   \setstretch{0.85} 
    \vspace{-3pt}
    \parbox{\linewidth}{\begin{footnotesize}\emph{Note}: Green indicates that the model fulfills this function; red indicates that it does not. Yellow means that there is some ambiguity; for instance, the IAEA only refers violations to the Security Council which then potentially takes action, a process that could be counted as enforcement. Similarly, tracking of key AI inputs could be part of the IAIO model but is optional. In the case of CERN, despite its civilian focus, the research could be classified as dual-use to a degree. These institutions were chosen for comparison because they represent commonly discussed models for international AI governance.\protect\footnotemark\ The IAIO is based on the ICAO, IMO, and FATF models, and thus these are not listed because they share similar characteristics.\end{footnotesize}\\}
    \caption{\textbf{Features of institutional analogies for AI governance models.}}     
    \label{table1final}
\end{table}
\footnotetext{See Ho et al., ``International Institutions for Advanced AI.''}

Firstly, proposals\footnote{Altman, Brockman, and Sutskever, ``Governance of Superintelligence''; Ho et al., ``International Institutions for Advanced AI.''} exist to centralize monitoring and inspection of all AI activities in an international institution (akin to the IAEA for nuclear technologies), referring violations to the UN Security Council. While such an institution could act as an independent reviewer in the context of AI governance, inspections along the lines of the IAEA model, which are used to verify the representations only of the non-nuclear weapons states in the context of the Nuclear Nonproliferation Treaty, might prove more challenging in an AI context. Furthermore, the process of referring violations to the UN Security Council is highly politicized. Given the rapid development in the field of AI, faster responses to compliance issues are advisable. In addition, AI development today is led predominantly by universities and the private sector, unlike nuclear technology, which was initially developed by states. The IAIO model enables agile governance of firms and governments by focusing on jurisdictional monitoring and state enforcement capabilities alongside agreed-upon minimum safety standards for the global industry. 

Secondly, there are proposals without monitoring and compliance components. For example, an Intergovernmental Panel on Climate Change (IPCC) equivalent for AI\footnote{Martin Rees, Shivaji Sondhi, and K VijayRaghavan, ``G20 Must Set up an International Panel on Technological Change,'' \emph{Hindustan Times}, March 19, 2023, \url{https://www.hindustantimes.com/opinion/g20-must-set-up-an-international-panel-on-technological-change-101679237287848.html}.} could exist to centralize information gathering about the state of AI into an international institution and to develop a global consensus around the risks from AI. Another example would be an international organization that centralizes AI capabilities research, or AI safety research, similar to the European Organization for Nuclear Research (CERN) for particle physics. Such proposals can have the goal either to centralize AI capabilities research in order to mitigate competitive risk-taking, or to increase AI safety research in order to investigate risks from AI along with potential solutions, especially those that are not investigated by for-profit organizations. These proposals do not focus on governing the respective technologies and can thus be complementary to an IAIO model by informing, for example, minimum safety standards based on conducted or synthesized research.

Finally, we could imagine club approaches, which are closest in spirit to the civilian governance model described above. Instead of a global standards regime, a group of aligned countries might set their own standards together. This would have the benefit of ameliorating the proliferation concerns: aligned states would be more willing to share information with each other, and more trusting of each other’s judgements when they are not willing to share. Furthermore, the fewer actors involved in such a governance regime, the more rapidly it could be set up. However, it would have the drawback of undermining legitimacy and standards compliance among other states. Note that the club and global regime models are not mutually exclusive. It may be that one set of standards can be developed and applied globally, while other, potentially more restrictive standards, are enforced among aligned states, for example through regional standards bodies and trade agreements.

\section{Conclusion}

AI presents a rapidly evolving international governance challenge. The technical advances of the last few years have led to remarkable new capabilities and disquieting realizations about how society could be harmed by this technology. While governance conversations have begun in many states, domestic governance alone will not be sufficient. International governance is needed to address both the highly international AI industry and the global reach of AI’s effects.

This report examines some key trade-offs in the international governance of civilian AI and describes one approach in detail. Civilian AI is the focus of these efforts since governance of that sector appears both feasible and urgently needed. 

We describe an international governance system that can ensure that AI regulation is standardized across participating states. It is composed of three key parts:

\begin{enumerate}
    \item A \textbf{standard setting body} codifies requirements for the specific behaviors of domestic AI regulation agencies. 
    \item A \textbf{jurisdictional certification body} certifies states if they achieve and maintain full compliance with the international standards. This body also provides assistance to states who request support in developing regulatory regimes to implement standards.
    \item \textbf{Domestic laws} give force to these certification decisions by requiring that trade in AI goods or precursors be conducted only with certified states.
\end{enumerate}

One key purpose of this group of institutions is to mitigate the most dangerous forms of competition among firms and states. In a competitive environment, a team implementing a safe, controllable, and socially acceptable AI may be preempted by a team that deploys an AI system that lacks one or more of these features. Domestic regulation can mitigate these concerns, as firms within the same jurisdiction will be subject to rules ensuring that a minimum standard is met for all AI products. However, unless similarly restrained, competition among states could lead to a ``race to the bottom'' on regulatory strength. The governance regime described above addresses this concern. It also ensures that the latest best practices in standards propagate globally and are enforced promptly by domestic authorities. Standards would be set by a competent international body; participating states would abide by the standards in order to trade with each other; all firms in these jurisdictions would face consistent regulatory expectations; and all civilians and states would live in a safer world.

This system is also designed to minimize the potential for the international governance system to serve as a vector for proliferation. Since domestic agencies are responsible for monitoring firms, these agencies can serve as a ``firewall'' that minimizes the unnecessary or unauthorized flow of information to international authorities. Proliferation will remain a key challenge for the regime, but the overall design of this approach should make it easier to control proliferation compared to more centralized governance approaches. 

International standards for civilian AI must evolve if they are to keep pace with the rapidly changing technological frontier. The governance regime described here can be designed to be agile and iterative, perhaps particularly if it is formed—like the International Organization for Standards and the International Accounting Standards Board—with a private or public-private partnership governance structure. The standard setting body can be tasked with regularly revising its regulations, the auditing body can update standards rapidly, and international standards can mandate domestic regulatory approaches that can respond with similar speed. Through these mechanisms, the regime is designed to evolve as it learns from its own prior iterations and keeps pace with a swiftly changing technology.

This international governance approach is broadly similar to those already in place in other industries, including civil aviation and shipping. The success of these analogous governance regimes lends credence to the idea that similar regimes are feasible for civilian AI. Under the ICAO regime, for instance, the number of worldwide civil aviation accidents decreased from 41 in 1944, the year the organization was founded, to 23 in 2019, despite a many thousandfold increase in passengers carried.\footnote{See ``Statistics $>$ By Period,'' Aviation Safety Network, n.d., \url{http://aviation-safety.net/statistics/period/stats.php} and International Civil Aviation Organization, ``Effects of Novel Coronavirus (COVID-19) on Civil Aviation: Economic Impact Analysis'' (Montréal, Canada, April 27, 2023), \url{https://www.icao.int/sustainability/Documents/Covid-19/ICAO\_coronavirus\_Econ\_Impact.pdf}.} Under the FATF regime, 76$\%$ of countries came into compliance with the FATF’s 40 recommendations in 2022 compared to 36$\%$ in 2012.\footnote{Financial Action Task Force, ``Report on the State of Effectiveness and Compliance with the FATF Standards,'' 2022, \url{https://www.fatf-gafi.org/en/publications/Fatfgeneral/Effectiveness-compliance-standards.html}. Note, however, that trends in money laundering, and thus the organization’s effectiveness are inherently hard to assess. For both sides of a debate on these issues, see Mark T. Nance, ``The Regime That FATF Built: An Introduction to the Financial Action Task Force,'' \emph{Crime, Law and Social Change} 69, no. 2 (2018): 109–29, \url{https://doi.org/10.1007/s10611-017-9747-6}.} 

The international governance of AI may require multiple interacting and even overlapping regimes. Military AI may end up being governed by very different agreements than civilian AI does. Furthermore, regional blocs or clubs of nations may place additional requirements on their firms that go beyond the global standard. The regime described in this report is compatible with many of these alternatives. 

These considerations lead to a series of near-term recommendations. States, industry, and civil society should endeavor to develop consensus on minimum regulatory standards for civilian AI. States around the world should be encouraged to create domestic regulatory capacities for AI and use a global summit to initiate a process for setting up an international civilian AI regulatory regime. The summit should be used to develop consensus on milestones for decision-making about the regime, and the milestone process should complete within six months of the summit.\footnote{Negotiating the Chicago Convention, which established the ICAO, took place between 52 governments over an intense month following US President Roosevelt’s invitation in September 1944. The Convention was signed on December 7th. See Jeffrey N Shane, ``Diplomacy and Drama: The Making of the Chicago Convention,'' \emph{Air \& Space Lawyer} 32, no. 4 (2019), \url{https://www.americanbar.org/content/dam/aba/publications/air\_space\_lawyer/Winter2019/as\_shane.pdf}. The formation of the FATF by the G7 was similarly rapid following the decision of the French and US governments to support it. See Mark Pieth and Gemma Aiolfi, eds., \emph{A Comparative Guide to Anti-Money Laundering: A Critical Analysis of Systems in Singapore, Switzerland, the UK and the USA} (Cheltenham, UK: Edward Elgar Publishing, 2004), pp. 8–9. Of course, many negotiations take much longer, particularly when groups of countries have divergent interests.} A core group of experts and frontier states can manage the milestones process with input from all UN states as well as non-governmental stakeholders, such as relevant NGOs, unions, and consumer groups. At the same time, efforts should be made to build broad public support for the proposed institution. The institution’s board should be structured to respect the interests of essential actors and mitigate against the organization being employed for political ends outside of its mandate. It should contain representatives from the technical and civil society AI governance communities, frontier AI states, and non-frontier AI states. Special care will be needed to prevent states from attempting to use a monitoring organization to gain access to frontier lab technologies.\footnote{Though it is slightly outside the scope of the paper, we believe the regime would be strengthened if the international community also takes steps to begin tracking all AI-specialized computing hardware. This would facilitate a variety of future governance efforts.}

Due to its complexity and potential, advanced AI may be very difficult to govern. Nonetheless, governance tools are available to address this challenge in the civilian domain. While much more work is needed in order to fill out the details, it is already possible to glimpse the outline of an interlocking regulatory landscape that can protect global society from the harmful aspects of this extraordinary and unprecedented technology.

\appendix

\section{AI Product and Precursor Trade Restrictions’ Compliance with International Trade Law}

Prima facie, import or export controls of AI products and precursors are compatible with international trade law.\footnote{While it is debatable whether international trade law applies to artificial intelligence given the lack of specific agreements or commitments in this area, this argument assumes that international trade law does apply to AI products and precursors in line with precedents from the WTO Appellate Body concerning emerging technologies. See Anupam Chander, ``Artificial Intelligence and Trade,'' in \emph{Big Data and Global Trade Law}, ed. Mira Burri (Cambridge, UK: Cambridge University Press, 2021), 115–27, \url{https://doi.org/10.1017/9781108919234.008}; World Trade Organization, ``DS363: China — Measures Affecting Trading Rights and Distribution Services for Certain Publications and Audiovisual Entertainment Products,'' December 21, 2019, \url{https://www.wto.org/english/tratop\_e/dispu\_e/cases\_e/ds363\_e.htm}, para 396.} The General Agreement on Tariffs and Trade (GATT) and the General Agreement on Trade in Services (GATS) indicate that, as long as actions taken by governments are not an arbitrary or unjustifiable discrimination or a ``disguised restriction on international trade,''\footnote{``The General Agreement on Tariffs and Trade (GATT),'' World Trade Organization, n.d., \url{https://www.wto.org/english/docs\_e/legal\_e/gatt47\_01\_e.htm}, arts I and III.} State Parties can take measures that are necessary to ensure safety and protect life or health, among other grounds.\footnote{``GATT,'' arts I, III, XX(b); ``General Agreement on Trade in Services (GATS),'' World Trade Organization, n.d., \url{https://www.wto.org/english/docs\_e/legal\_e/26-gats\_01\_e.htm}, art XIV. See also ``Agreement on the Application of Sanitary and Phytosanitary Measures (SPS Agreement),'' World Trade Organization, n.d., \url{https://www.wto.org/english/tratop\_e/sps\_e/spsagr\_e.htm}, art 1.1.} Furthermore, the GATT, GATS, and the Agreement on Trade-Related Aspects of Intellectual Property Rights (TRIPS) establish security exceptions indicating that nothing in those treaties can be construed to prevent a State Party ``from taking any action which \emph{it considers necessary }for the protection of its essential security interests,'' including measures taken to comply with their international peace and security obligations under the UN Charter.\footnote{``GATT,'' arts XXI(b)(iii), XXI(c); ``GATS,'' art XIV bis; ``Agreement on Trade-Related Aspects of Intellectual Property Rights (TRIPS),'' World Trade Organization, n.d., \url{https://www.wto.org/english/docs\_e/legal\_e/31bis\_trips\_01\_e.htm}, art 73. Emphasis added.} Trade restrictions under these exceptions would usually apply to a final product but can arguably also apply to process and production methods.\footnote{See Andreas R. Ziegler and David Sifonios, ``The Assessment of Environmental Risks and the Regulation of Process and Production Methods (PPMs) in International Trade Law,'' in \emph{Risk and the Regulation of Uncertainty in International Law}, ed. Mónika Ambrus, Rosemary Rayfuse, and Wouter Werner (Oxford, UK: Oxford University Press, 2017), 219–36.}

The measures proposed in this report would also constitute legitimate technical trade barriers. Under the Technical Barriers to Trade Agreement (TBT), countries may enact legal requirements to ensure that imported or exported products comply with national security requirements, to guarantee that they are safe, or to prevent deceptive practices, among other legitimate objectives.\footnote{``Agreement on Technical Barriers to Trade (TBT),'' World Trade Organization, n.d., \url{https://www.wto.org/english/docs\_e/legal\_e/17-tbt\_e.htm}, art 2.2.} Given the temptation to accord a more favorable treatment to national products or to products from certain countries, the TBT relies heavily on international standards as the leading basis for the adoption of technical barriers.\footnote{``TBT,'' arts 2.4, 2.6.} Such international standards could be set by an International AI Organization.

Notably, any trade restrictions applied by states under the model proposed in this paper may need to pass a ``necessity test'' at the World Trade Organization (WTO). This necessity test takes into account four requirements:\footnote{See World Trade Organization, ``DS161: Korea — Measures Affecting Imports of Fresh, Chilled and Frozen Beef,'' January 10, 2001, \url{https://www.wto.org/english/tratop\_e/dispu\_e/cases\_e/ds161\_e.htm}, para. 164; World Trade Organization, ``DS285: United States — Measures Affecting the Cross-Border Supply of Gambling and Betting Services,'' April 20, 2005, \url{https://www.wto.org/english/tratop\_e/dispu\_e/cases\_e/ds285\_e.htm}, paras. 304–307.} 

\begin{enumerate}
	\item  The relative importance of the protected public interest(s) pursued by a measure; 

	\item The contested measure’s contribution to the achievement of the objective that is being pursued;

	\item The trade restrictiveness of the measure; and 

	\item A determination of whether, in the light of importance of the interests at issue, a less trade restrictive alternative is ``reasonably available.''

\end{enumerate}
While WTO panels have not always interpreted these elements consistently, it is safe to assume that, in light of the multiple risks from AI that have been highlighted in this report, the first of the factors listed above would be met. Additionally, the vast importance of protecting people’s lives and wellbeing from those risks would weigh heavily in favor of justifying a measure’s degree of restrictiveness. Meeting the second and fourth requirements of the necessity test would depend on the specific design of the import and export controls. However, taking into account that controls would be in line with internationally agreed standards, based on a common understanding of the objectives being pursued and the associated costs, it seems likely that controls would meet the necessity test as long as they are effective at mitigating risks from AI.

\clearpage

\section*{Bibliography}
\addcontentsline{toc}{section}{Bibliography}

\mybibitem Abbott, Kenneth W., and Duncan Snidal. ``The Governance Triangle:
Regulatory Standards Institutions and the Shadow of the State.'' In
\emph{The Politics of Global Regulation}, edited by Walter Mattli and
Ngaire Woods, 44--88. Princeton University Press, 2009.
\url{https://doi.org/10.1515/9781400830732.44}.

\mybibitem ABP News Bureau. ``BRICS Nations Call For Effective Global Framework On
AI, Emphasise On Ethical Development.'' ABP News Live, June 2, 2023.
\url{https://news.abplive.com/technology/ai-brics-nations-call-for-effective-global-framework-on-artificial-intelligence-emphasise-on-ethical-development-1606406}.

\mybibitem AI Now Institute. ``2023 Landscape Executive Summary,'' 2023.
\url{https://ainowinstitute.org/general/2023-landscape-executive-summary}.

\mybibitem Altman, Sam. ``Machine Intelligence, Part 1,'' February 25, 2015.
\url{https://blog.samaltman.com/machine-intelligence-part-1}.

\mybibitem Altman, Sam, Greg Brockman, and Ilya Sutskever. ``Governance of
Superintelligence.'' OpenAI, May 22, 2023.
\url{https://openai.com/blog/governance-of-superintelligence}.

\mybibitem Anderljung, Markus, Joslyn Barnhart, Anton Korinek, Jade Leung, Cullen
O'Keefe, Jess Whittlestone, et al. ``Frontier AI Regulation: Managing
Emerging Risks to Public Safety.'' arXiv, 2023. arXiv:2307.03718.

\mybibitem ARC Evals. ``Update on ARC's Recent Eval Efforts,'' March 17, 2023.
\url{https://evals.alignment.org/blog/2023-03-18-update-on-recent-evals/}.

\mybibitem \emph{Artificial Intelligence: Opportunities and Risks for International
Peace and Security - Security Council, 9381st Meeting}. United Nations
Security Council, 2023.
\url{https://media.un.org/en/asset/k1j/k1ji81po8p}.

\mybibitem Askell, Amanda, Miles Brundage, and Gillian Hadfield. ``The Role of
Cooperation in Responsible AI Development.'' arXiv, 2019.
arXiv:1907.04534.

\mybibitem Aviation Safety Network. ``Statistics \textgreater{} By Period,'' n.d.
\url{http://aviation-safety.net/statistics/period/stats.php}.

\mybibitem Aytbaev, Bulat, Dmitry Grigoriev, Vladislav Lavrenchuk, and Noah C
Mayhew. ``Don't Let Nuclear Accidents Scare You Away from Nuclear
Power.'' \emph{Bulletin of the Atomic Scientists}, August 31, 2020.
\url{https://thebulletin.org/2020/08/dont-let-nuclear-accidents-scare-you-away-from-nuclear-power/}.

\mybibitem Baker, Mauricio. ``Nuclear Arms Control Verification and Lessons for AI
Treaties.'' arXiv, 2023. arXiv:2304.04123.

\mybibitem Barratt-Brown, Elizabeth P. ``Building a Monitoring and Compliance
Regime Under the Montreal Protocol.'' \emph{Yale Journal of
International Law} 16 (1991): 519--70.
\url{https://openyls.law.yale.edu/handle/20.500.13051/6255}.

\mybibitem Bengio, Yoshua. ``How Rogue AIs May Arise,'' May 22, 2023.
\url{https://yoshuabengio.org/2023/05/22/how-rogue-ais-may-arise/}.

\mybibitem Bengio, Yoshua. ``Slowing Down Development of AI Systems Passing the
Turing Test,'' April 5, 2023.
\url{https://yoshuabengio.org/2023/04/05/slowing-down-development-of-ai-systems-passing-the-turing-test/}.

\mybibitem Bluth, Christoph, Matthew Kroenig, Rensselaer Lee, William C. Sailor,
and Matthew Fuhrmann. ``Civilian Nuclear Cooperation and the
Proliferation of Nuclear Weapons.'' \emph{International Security} 35,
no. 1 (2010): 184--200. \url{https://www.jstor.org/stable/40784651}.

\mybibitem Bohdal, Ondrej, Timothy Hospedales, Philip H. S. Torr, and Fazl Barez.
``Fairness in AI and Its Long-Term Implications on Society.'' arXiv,
2023. arXiv:2304.09826..

\mybibitem Bommasani, Rishi, Drew A. Hudson, Ehsan Adeli, Russ Altman, Simran
Arora, Sydney von Arx, Michael S. Bernstein, et al. ``On the
Opportunities and Risks of Foundation Models.'' arXiv, 2022.
arXiv:2108.07258.

\mybibitem Bradford, Anu. The Brussels Effect: How the European Union Rules the
World. Oxford University Press, 2020.

\mybibitem Brundage, Miles, Shahar Avin, Jack Clark, Helen Toner, Peter Eckersley,
Ben Garfinkel, Allan Dafoe, et al. ``The Malicious Use of Artificial
Intelligence: Forecasting, Prevention, and Mitigation.'' arXiv, 2018.
arXiv:1802.07228.

\mybibitem Brundage, Miles, Shahar Avin, Jasmine Wang, Haydn Belfield, Gretchen
Krueger, Gillian Hadfield, Heidy Khlaaf, et al. ``Toward Trustworthy AI
Development: Mechanisms for Supporting Verifiable Claims.'' arXiv, 2020.
arXiv:2004.07213.

\mybibitem Bucknall, Ben, Toby Shevlane, and Robert Trager. ``Structured Access for
Third-Party Safety Research on Frontier AI Models Investigating
Researchers' Model Access Requirements.'' Working Paper, n.d.

\mybibitem Büthe, Tim, and Walter Mattli. \emph{The New Global Rulers: The
Privatization of Regulation in the World Economy.} Princeton, NJ:
Princeton University Press, 2011.

\mybibitem Carnegie, Allison, and Austin Carson. ``The Disclosure Dilemma: Nuclear
Intelligence and International Organizations.'' \emph{American Journal
of Political Science} 63, no. 2 (April 2019): 269--85.
\url{https://doi.org/10.1111/ajps.12426}.

\mybibitem Center for AI Safety. ``Statement on AI Risk,'' May 30, 2023.
https://www.safe.ai/statement-on-ai-risk.

\mybibitem Chander, Anupam. ``Artificial Intelligence and Trade.'' In \emph{Big
Data and Global Trade Law}, edited by Mira Burri, 115--27. Cambridge,
UK: Cambridge University Press, 2021.
\url{https://doi.org/10.1017/9781108919234.008}.

\mybibitem Chow, Andrew. ``How ChatGPT Managed to Grow Faster Than TikTok or
Instagram.'' \emph{Time}, February 8, 2023.
\url{https://time.com/6253615/chatgpt-fastest-growing/}.

\mybibitem ``Civil Aviation Law of the People's Republic of China.'' General
Administration of Civil Aviation of China, 1995.
\url{http://www.caac.gov.cn/en/ZCFG/MHFL/201509/P020150901511659239730.pdf}.

\mybibitem Coalition for Content Provenance and Authenticity. ``Overview.''
Accessed July 14, 2023. \url{https://c2pa.org/}.

\mybibitem Cottier, Ben. ``Trends in the Dollar Training Cost of Machine Learning
Systems.'' Epoch, 2023.
\url{https://epochai.org/blog/trends-in-the-dollar-training-cost-of-machine-learning-systems}.

\mybibitem Deloitte. ``Supply Chain Model: Chain of Custody,'' n.d.
\url{https://www2.deloitte.com/nl/nl/pages/sustainability/articles/chain-of-custody.html}.

\mybibitem Epoch. ``Parameter, Compute and Data Trends in Machine Learning,'' 2022.
\url{https://epochai.org/data/pcd}.

\mybibitem Erdil, Ege, and Tamay Besiroglu. ``Algorithmic Progress in Computer
Vision.'' arXiv, 2023. arXiv:2212.05153.

\mybibitem ``Evolving ICAO's Universal Safety Oversight Audit Programme: The
Continuous Monitoring Approach.'' \emph{ICAO Journal} 65, no. 4 (2010):
24--25.
\url{https://www.icao.int/safety/CMAForum/Shared\%20Documents/6504_en-1.pdf}.

\mybibitem FAMI QS. ``Accreditation Bodies.'' Accessed June 22, 2023.
\url{https://fami-qs.org/certified-organisations/accreditation-bodies/}.

\mybibitem Federal Aviation Administration. ``International Aviation Safety
Assessment (IASA) Program.'' Accessed June 22, 2023.
\url{https://www.faa.gov/sites/faa.gov/files/about/initiatives/iasa/FAA_Initiatives_IASA.pdf}.

\mybibitem Federal Aviation Administration. ``International Aviation Safety
Assessment (IASA) Program,'' n.d.
\url{https://www.faa.gov/about/initiatives/iasa}.

\mybibitem Financial Action Task Force. ``\,`Black and Grey' Lists,'' n.d.
\url{https://www.fatf-gafi.org/en/countries/black-and-grey-lists.html}.

\mybibitem Financial Action Task Force. ``Methodology for Assessing Compliance with
the FATF Recommendations and the Effectiveness of AML/CFT Systems,''
2023.
\url{https://www.fatf-gafi.org/en/publications/Mutualevaluations/Fatf-methodology.html}.

\mybibitem Financial Action Task Force. ``Mutual Evaluations.'' Accessed July 13,
2023. \url{https://www.fatf-gafi.org/en/topics/mutual-evaluations.html}.

\mybibitem Financial Action Task Force. ``Report on the State of Effectiveness and
Compliance with the FATF Standards,'' 2022.
\url{https://www.fatf-gafi.org/en/publications/Fatfgeneral/Effectiveness-compliance-standards.html}.

\mybibitem Forest Stewardship Council. ``Certification System.'' Accessed June 22,
2023. \url{https://connect.fsc.org/certification/certification-system}.

\mybibitem Forest Stewardship Council. ``Chain of Custody Certification.'' Accessed
July 13, 2023. \url{https://fsc.org/en/chain-of-custody-certification}.

\mybibitem Forest Stewardship Council UK. ``FSC-Accredited Certification Bodies.''
Accessed June 22, 2023.
\url{https://uk.fsc.org/fsc-accredited-certification-bodies}.

\mybibitem Forest Stewardship Council UK. ``Governance,'' n.d.
https://uk.fsc.org/governance.

\mybibitem Fuhrmann, Matthew. \emph{Atomic Assistance: How ``Atoms for Peace''
Programs Cause Nuclear Insecurity}. Ithaca, NY: Cornell University
Press, 2012.

\mybibitem Future of Life Institute. ``Pause Giant AI Experiments: An Open
Letter,'' March 22, 2023.
\url{https://futureoflife.org/open-letter/pause-giant-ai-experiments/}.

\mybibitem Ganguli, Deep, Danny Hernandez, Liane Lovitt, Amanda Askell, Yuntao Bai,
Anna Chen, Tom Conerly, et al. ``Predictability and Surprise in Large
Generative Models.'' In Proceedings of the 2022 ACM Conference on
Fairness, Accountability, and Transparency, 1747--64. FAccT '22. ACM,
2022. \url{https://doi.org/10.1145/3531146.3533229}.

\mybibitem Gibbs, Samuel. ``Elon Musk: Artificial Intelligence Is Our Biggest
Existential Threat.'' \emph{The Guardian}, October 27, 2014.
\url{https://www.theguardian.com/technology/2014/oct/27/elon-musk-artificial-intelligence-ai-biggest-existential-threat}.

\mybibitem Goldstein, Josh A., Girish Sastry, Micah Musser, Renee DiResta, Matthew
Gentzel, and Katerina Sedova. ``Generative Language Models and Automated
Influence Operations: Emerging Threats and Potential Mitigations.''
arXiv, 2023. arXiv:2301.04246.

\mybibitem \emph{Google CEO Calls for Global AI Regulation}. 60 Minutes, April 16,
2023. \url{https://www.youtube.com/watch?v=aNsmr-tvQhA}.

\mybibitem Google. ``A Policy Agenda for Responsible AI Progress: Opportunity,
Responsibility, Security.'' May 19, 2023.
\url{https://blog.google/technology/ai/a-policy-agenda-for-responsible-ai-progress-opportunity-responsibility-security/}.

\mybibitem GOV.UK. ``PM Meeting with Leading CEOs in AI: 24 May 2023,'' May 24,
2023.
\url{https://www.gov.uk/government/news/pm-meeting-with-leading-ceos-in-ai-24-may-2023}.

\mybibitem GOV.UK. ``UK to Host First Global Summit on Artificial Intelligence,''
June 7, 2023.
\url{https://www.gov.uk/government/news/uk-to-host-first-global-summit-on-artificial-intelligence}.

\mybibitem Guterres, António. ``Secretary-General Urges Broad Engagement from All
Stakeholders towards United Nations Code of Conduct for Information
Integrity on Digital Platforms'' United Nations, June 12, 2023.
\url{https://press.un.org/en/2023/sgsm21832.doc.htm}.

\mybibitem Guterres, António. ``Secretary-General's Remarks to the Security Council
on Artificial Intelligence.'' United Nations Security Council, July 18,
2023.
\url{https://www.un.org/sg/en/content/sg/speeches/2023-07-18/secretary-generals-remarks-the-security-council-artificial-intelligence}.

\mybibitem Hazell, Julian. ``Large Language Models Can Be Used To Effectively Scale
Spear Phishing Campaigns.'' arXiv, 2023. arXiv:2305.06972.

\mybibitem Henighan, Tom, Jared Kaplan, Mor Katz, Mark Chen, Christopher Hesse,
Jacob Jackson, Heewoo Jun, et al. ``Scaling Laws for Autoregressive
Generative Modeling.'' arXiv, 2020. arXiv:2010.14701.

\mybibitem Hestness, Joel, Sharan Narang, Newsha Ardalani, Gregory Diamos, Heewoo
Jun, Hassan Kianinejad, Md Mostofa Ali Patwary, Yang Yang, and Yanqi
Zhou. ``Deep Learning Scaling Is Predictable, Empirically.'' arXiv,
2017. arXiv:1712.00409.

\mybibitem Ho, Lewis, Joslyn Barnhart, Robert Trager, Yoshua Bengio, Miles
Brundage, Allison Carnegie, Rumman Chowdhury, et al. ``International
Institutions for Advanced AI.'' arXiv, 2023. arXiv:2307.04699.

\mybibitem Hoffmann, Jordan, Sebastian Borgeaud, Arthur Mensch, Elena Buchatskaya,
Trevor Cai, Eliza Rutherford, Diego de Las Casas, et al. ``Training
Compute-Optimal Large Language Models.'' arXiv, 2022. arXiv:2203.15556.

\mybibitem Hoffman, Steven J., Prativa Baral, Susan Rogers Van Katwyk, Lathika
Sritharan, Matthew Hughsam, Harkanwal Randhawa, Gigi Lin, et al.
``International Treaties Have Mostly Failed to Produce Their Intended
Effects.'' \emph{Proceedings of the National Academy of Sciences} 119,
no. 32 (August 9, 2022): e2122854119.
\url{https://doi.org/10.1073/pnas.2122854119}.

\mybibitem Huang, Seaton, Helen Toner, Zac Haluza, and Rogier Creemers, trans.
``Translation: Measures for the Management of Generative Artificial
Intelligence Services (Draft for Comment) -- April 2023,'' April 12,
2023.
\url{https://digichina.stanford.edu/work/translation-measures-for-the-management-of-generative-artificial-intelligence-services-draft-for-comment-april-2023/}.

\mybibitem International Atomic Energy Agency, ``The Statute of the IAEA'' (n.d.),
\url{https://www.iaea.org/about/statute}.

\mybibitem International Civil Aviation Organization. ``Another `No Country Left
Behind' Success: A Significant Safety Concern Resolved!,'' n.d.
\url{https://www.icao.int/EURNAT/Pages/news_articles/NoCountryLeftBehind-success.aspx}.

\mybibitem International Civil Aviation Organization. ``Effects of Novel
Coronavirus (COVID-19) on Civil Aviation: Economic Impact Analysis.''
Montréal, Canada, April 27, 2023.
\url{https://www.icao.int/sustainability/Documents/Covid-19/ICAO_coronavirus_Econ_Impact.pdf}.

\mybibitem International Civil Aviation Organization. ``Frequently Asked Questions
about USOAP,'' n.d.
\url{https://www.icao.int/safety/CMAForum/Pages/FAQ.aspx}.

\mybibitem International Civil Aviation Organization. ``Personnel Licensing FAQ,''
n.d.
\url{https://www.icao.int/safety/airnavigation/pages/peltrgfaq.aspx}.

\mybibitem International Electrotechnical Commission. ``How the Global IEC
Conformity Assessment Systems Operate: A Network of Trust.'' Accessed
June 22, 2023.
\url{https://www.iec.ch/conformity-assessment/how-global-iec-ca-systems-operate}.

\mybibitem International Maritime Organization. ``Framework and Procedures for the
IMO Member State Audit Scheme,'' December 5, 2013.
\url{https://wwwcdn.imo.org/localresources/en/OurWork/MSAS/Documents/MSAS/Basic\%20documents/A.1067(28)\%20Framework\%20and\%20Procedures.pdf}.

\mybibitem International Maritime Organization. ``Frequently Asked Questions on
Maritime Security.'' Accessed June 9, 2023.
\url{https://www.imo.org/en/OurWork/Security/Pages/FAQ.aspx}.

\mybibitem Janjeva, Ardi, Nikhil Mulani, Rosamund Powell, Jess Whittlestone, and
Shahar Avin. ``Strengthening Resilience to AI Risk: A Guide for UK
Policymakers.'' Centre for Emerging Technology and Security, 2023.
\url{https://www.longtermresilience.org/post/paper-launch-strengthening-resilience-to-ai-risk-a-guide-for-uk-policymakers}.

\mybibitem Jannai, Daniel, Amos Meron, Barak Lenz, Yoav Levine, and Yoav Shoham.
``Human or Not? A Gamified Approach to the Turing Test.'' arXiv, 2023.
arXiv:2305.20010.

\mybibitem Jumper, John, Richard Evans, Alexander Pritzel, Tim Green, Michael
Figurnov, Olaf Ronneberger, Kathryn Tunyasuvunakool, et al. ``Highly
Accurate Protein Structure Prediction with AlphaFold.'' \emph{Nature}
596, no. 7873 (2021): 583--89.
\url{https://doi.org/10.1038/s41586-021-03819-2}.

\mybibitem Kaplan, Jared, Sam McCandlish, Tom Henighan, Tom B. Brown, Benjamin
Chess, Rewon Child, Scott Gray, Alec Radford, Jeffrey Wu, and Dario
Amodei. ``Scaling Laws for Neural Language Models.'' arXiv, 2020.
arXiv:2001.08361.

\mybibitem Khan, Saif M., and Alexander Mann. ``AI Chips: What They Are and Why
They Matter.'' Center for Security and Emerging Technology, 2020.
\url{https://doi.org/10.51593/20190014}.

\mybibitem Küspert, Sabrina, Nicolas Moës, and Connor Dunlop. ``The
Value\hspace{0pt}\hspace{0pt}\hspace{0pt}
\hspace{0pt}\hspace{0pt}\hspace{0pt}Chain of General-Purpose AI.'' Ada
Lovelace Institute, 2023.
\url{https://www.adalovelaceinstitute.org/blog/value-chain-general-purpose-ai/}.

\mybibitem Lall, Ranjit. \emph{Making International Institutions Work: The Politics
of Performance}. Cambridge, UK: Cambridge University Press, 2023.
\url{https://doi.org/10.1017/9781009216265}.

\mybibitem Laplante, Phil, and Ben Amaba. ``Artificial Intelligence in Critical
Infrastructure Systems.'' \emph{Computer} 54, no. 10 (2021): 14--24.
\url{https://doi.org/10.1109/MC.2021.3055892}.

\mybibitem Li, Zihao. ``Why the European AI Act Transparency Obligation Is
Insufficient.'' \emph{Nature Machine Intelligence} 5, no. 6 (2023):
559--60. \url{https://doi.org/10.1038/s42256-023-00672-y}.

\mybibitem Manheim, Karl, and Lyric Kaplan. ``Artificial Intelligence: Risks to
Privacy and Democracy.'' \emph{Yale Journal of Law and Technology} 21
(2019): 106--88.
\url{https://yjolt.org/artificial-intelligence-risks-privacy-and-democracy}.

\mybibitem Martínez, Eric. ``Re-Evaluating GPT-4's Bar Exam Performance.'' SSRN,
2023. \url{https://doi.org/10.2139/ssrn.4441311}.

\mybibitem Maslej, Nestor, Loredana Fattorini, Erik Brynjolfsson, John Etchemendy,
Katrina Ligett, Terah Lyons, James Manyika, et al. ``Artificial
Intelligence Index Report 2023.'' Institute for Human Centered AI, 2023.
\url{https://aiindex.stanford.edu/wp-content/uploads/2023/04/HAI_AI-Index-Report_2023.pdf}.

\mybibitem Metz, Cade. ``\,`The Godfather of A.I.' Leaves Google and Warns of
Danger Ahead.'' \emph{The New York Times}, May 1, 2023.
\url{https://www.nytimes.com/2023/05/01/technology/ai-google-chatbot-engineer-quits-hinton.html}.

\mybibitem Microsoft. ``Governing AI: A Blueprint for the Future,'' 2023.
\url{https://query.prod.cms.rt.microsoft.com/cms/api/am/binary/RW14Gtw}.

\mybibitem Mitchell, Margaret, Simone Wu, Andrew Zaldivar, Parker Barnes, Lucy
Vasserman, Ben Hutchinson, Elena Spitzer, Inioluwa Deborah Raji, and
Timnit Gebru. ``Model Cards for Model Reporting.'' In \emph{Proceedings
of the Conference on Fairness, Accountability, and Transparency},
220--29. FAT* '19. ACM, 2019.
\url{https://doi.org/10.1145/3287560.3287596}.

\mybibitem Mökander, Jakob, Jonas Schuett, Hannah Rose Kirk, and Luciano Floridi.
``Auditing Large Language Models: A Three-Layered Approach.'' \emph{AI
and Ethics}, 2023. \url{https://doi.org/10.1007/s43681-023-00289-2}.

\mybibitem Morgan, Forrest E., Benjamin Boudreaux, Andrew J. Lohn, Mark Ashby,
Christian Curriden, Kelly Klima, and Derek Grossman. ``Military
Applications of Artificial Intelligence: Ethical Concerns in an
Uncertain World.'' RAND Corporation, 2020.
\url{https://www.rand.org/pubs/research_reports/RR3139-1.html}.

\mybibitem Nance, Mark T. ``The Regime That FATF Built: An Introduction to the
Financial Action Task Force.'' \emph{Crime, Law and Social Change} 69,
no. 2 (2018): 109--29. \url{https://doi.org/10.1007/s10611-017-9747-6}.

\mybibitem Newhouse, John. \emph{War and Peace in the Nuclear Age}. New York:
Alfred A. Knopf, 1988.

\mybibitem Ngo, Richard, Lawrence Chan, and Sören Mindermann. ``The Alignment
Problem from a Deep Learning Perspective.'' arXiv, February 22, 2023.
arXiv:2209.00626.

\mybibitem Nichols, Thomas M. \emph{No Use: Nuclear Weapons and U.S. National
Security}. Philadelphia: University of Pennsylvania Press, 2013.

\mybibitem OpenAI. ``GPT-4 Technical Report.'' arXiv, 2023. arXiv:2303.08774.

\mybibitem OpenAI. ``Our Approach to AI Safety,'' April 5, 2023.
\url{https://openai.com/blog/our-approach-to-ai-safety}.

\mybibitem ``OpenAI Evals.'' OpenAI, 2023. \url{https://github.com/openai/evals}.

\mybibitem Ord, Toby. ``Lessons from the Development of the Atomic Bomb.'' Centre
for the Governance of AI, 2022.
\url{https://www.governance.ai/research-paper/lessons-atomic-bomb-ord}.

\mybibitem ``Oversight of A.I.: Rules for Artificial Intelligence.'' May 16, 2023.
\url{https://www.judiciary.senate.gov/committee-activity/hearings/oversight-of-ai-rules-for-artificial-intelligence}.

\mybibitem Perez, Ethan, Sam Ringer, Kamilė Lukošiūtė, Karina Nguyen, Edwin Chen,
Scott Heiner, Craig Pettit, et al. ``Discovering Language Model
Behaviors with Model-Written Evaluations.'' arXiv, December 19, 2022.
arXiv:2212.09251.

\mybibitem Pieth, Mark, and Gemma Aiolfi, eds. \emph{A Comparative Guide to
Anti-Money Laundering: A Critical Analysis of Systems in Singapore,
Switzerland, the UK and the USA}. Cheltenham, UK: Edward Elgar
Publishing, 2004.

\mybibitem Proposal for a Regulation of the European Parliament and of the Council
Laying down Harmonised Rules on Artificial Intelligence (Artificial
Intelligence Act) and Amending Certain Union Legislative Acts (2021).
\url{https://eur-lex.europa.eu/legal-content/EN/TXT/?uri=celex\%3A52021PC0206}.

\mybibitem Raji, Inioluwa Deborah, Andrew Smart, Rebecca N. White, Margaret
Mitchell, Timnit Gebru, Ben Hutchinson, Jamila Smith-Loud, Daniel
Theron, and Parker Barnes. ``Closing the AI Accountability Gap: Defining
an End-to-End Framework for Internal Algorithmic Auditing.'' arXiv,
2020. arXiv:2001.00973.

\mybibitem Rawlinson, Kevin. ``Microsoft's Bill Gates Insists AI Is a Threat.''
\emph{BBC News}, January 29, 2015.
\url{https://www.bbc.co.uk/news/31047780}.

\mybibitem ``Regulation (EU) 2018/1139 of the European Parliament and of the
Council.'' European Union, July 4, 2018.
\url{https://eur-lex.europa.eu/legal-content/EN/TXT/?uri=CELEX\%3A32018R1139}.

\mybibitem Rees, Martin, Shivaji Sondhi, and K VijayRaghavan. ``G20 Must Set up an
International Panel on Technological Change.'' \emph{Hindustan Times},
March 19, 2023.
\url{https://www.hindustantimes.com/opinion/g20-must-set-up-an-international-panel-on-technological-change-101679237287848.html}.

\mybibitem \emph{Reuters}. ``Britain, U.S. to Work Together on AI Safety, Says
Sunak,'' June 8, 2023.
\url{https://www.reuters.com/technology/britain-us-work-together-ai-safety-says-sunak-2023-06-08/}.

\mybibitem Roehrlich, Elisabeth. Inspectors for Peace: A History of the
International Atomic Energy Agency. Baltimore: JHU Press, 2022.

\mybibitem Russell, Stuart. \emph{Human Compatible: Artificial Intelligence and the
Problem of Control}. Penguin, 2019.

\mybibitem Schaeffer, Rylan, Brando Miranda, and Sanmi Koyejo. ``Are Emergent
Abilities of Large Language Models a Mirage?'' arXiv, 2023.
arXiv:2304.15004.

\mybibitem Schuett, Jonas. ``Defining the Scope of AI Regulations.'' Legal
Priorities Project Working Paper No. 9. 2021.
\url{https://papers.ssrn.com/abstract=3453632}.

\mybibitem Sevilla, Jaime, Lennart Heim, Anson Ho, Tamay Besiroglu, Marius
Hobbhahn, and Pablo Villalobos. ``Compute Trends Across Three Eras of
Machine Learning.'' arXiv, 2022. arXiv:2202.05924.

\mybibitem Shane, Jeffrey N. ``Diplomacy and Drama: The Making of the Chicago
Convention.'' \emph{Air \& Space Lawyer} 32, no. 4 (2019).
\url{https://www.americanbar.org/content/dam/aba/publications/air_space_lawyer/Winter2019/as_shane.pdf}.

\mybibitem Shavit, Yonadav. ``What Does It Take to Catch a Chinchilla? Verifying
Rules on Large-Scale Neural Network Training Via Compute Monitoring.''
arXiv, 2023. arXiv:2303.11341.

\mybibitem Sheng, Ying, Lianmin Zheng, Binhang Yuan, Zhuohan Li, Max Ryabinin,
Daniel Y. Fu, Zhiqiang Xie, et al. ``FlexGen: High-Throughput Generative
Inference of Large Language Models with a Single GPU.'' arXiv, 2023.
arXiv:2303.06865.

\mybibitem Shevlane, Toby. ``Structured Access: An Emerging Paradigm for Safe AI
Deployment.'' arXiv, 2022. arXiv:2201.05159.

\mybibitem Shevlane, Toby, Sebastian Farquhar, Ben Garfinkel, Mary Phuong, Jess
Whittlestone, Jade Leung, Daniel Kokotajlo, et al. ``Model Evaluation
for Extreme Risks.'' arXiv, 2023. arXiv:2305.15324.

\mybibitem Siegmann, Charlotte, and Markus Anderljung. ``The Brussels Effect and
Artificial Intelligence: How EU Regulation Will Impact the Global AI
Market.'' arXiv, 2022. arXiv:2208.12645.

\mybibitem Simpson, Morgan, and Robert Trager. ``Cooperation in Safety-Critical
Industries: Lessons for AI from Aviation and Nuclear.'' Working Paper,
n.d.

\mybibitem Stafford, Eoghan, Robert Trager, and Allan Dafoe. ``Safety Not
Guaranteed: International Strategic Dynamics of Risky Technology
Races.'' Working Paper. Centre for the Governance of AI, 2022.
\url{https://www.governance.ai/research-paper/safety-not-guaranteed-international-strategic-dynamics-of-risky-technology-ra}.

\mybibitem Temtsin, Sharon, Diane Proudfoot, and Christoph Bartneck. ``A Bona Fide
Turing Test.'' In \emph{Proceedings of the 10th International Conference
on Human-Agent Interaction}, 250--52. HAI '22. ACM, 2022.
\url{https://doi.org/10.1145/3527188.3563918}.

\mybibitem The Australia Group. ``Control List of Dual-Use Biological Equipment and
Related Technology and Software.'' Accessed June 9, 2023.
\url{https://www.dfat.gov.au/publications/minisite/theaustraliagroupnet/site/en/dual_biological.html}.

\mybibitem The Australia Group. ``Dual-Use Chemical Manufacturing Facilities and
Equipment.'' Accessed June 9, 2023.
\url{https://www.dfat.gov.au/publications/minisite/theaustraliagroupnet/site/en/dual_chemicals.html}.

\mybibitem The Australia Group. ``Export Control List: Chemical Weapons
Precursors.'' Accessed June 9, 2023.
\url{https://www.dfat.gov.au/publications/minisite/theaustraliagroupnet/site/en/precursors.html}.

\mybibitem The White House. ``Readout of White House Meeting with CEOs on Advancing
Responsible Artificial Intelligence Innovation,'' May 4, 2023.
\url{https://www.whitehouse.gov/briefing-room/statements-releases/2023/05/04/readout-of-white-house-meeting-with-ceos-on-advancing-responsible-artificial-intelligence-innovation/}.

\mybibitem Toner, Helen, Zac Haluza, Yan Luo, Xuezi Dan, Matt Sheehan, Seaton
Huang, Kimball Chen, Rogier Creemers, Paul Triolo, and Caroline
Meinhardt. ``How Will China's Generative AI Regulations Shape the
Future? A DigiChina Forum.'' DigiChina, April 19, 2023.
\url{https://digichina.stanford.edu/work/how-will-chinas-generative-ai-regulations-shape-the-future-a-digichina-forum/}.

\mybibitem UK Department for Science, Innovation and Technology. ``AI Regulation: A
Pro-Innovation Approach,'' March 2023.
\url{https://www.gov.uk/government/publications/ai-regulation-a-pro-innovation-approach}.

\mybibitem United States Government Accountability Office. ``Aviation Security: TSA
Strengthened Foreign Airport Assessments and Air Carrier Inspections,
but Could Improve Analysis to Better Address Deficiencies,'' 2017.
\url{https://www.gao.gov/assets/gao-18-178.pdf}.

\mybibitem Urbina, Fabio, Filippa Lentzos, Cédric Invernizzi, and Sean Ekins.
``Dual Use of Artificial-Intelligence-Powered Drug Discovery.''
\emph{Nature Machine Intelligence} 4, no. 3 (2022): 189--91.
\url{https://doi.org/10.1038/s42256-022-00465-9}.

\mybibitem Verma, Pranshu. ``They Thought Loved Ones Were Calling for Help. It Was
an AI Scam.'' \emph{Washington Post}, March 5, 2023.
\url{https://www.washingtonpost.com/technology/2023/03/05/ai-voice-scam/}.

\mybibitem Villalobos, Pablo. ``Scaling Laws Literature Review.'' Epoch, 2023.
\url{https://epochai.org/blog/scaling-laws-literature-review}.

\mybibitem Wei, Jason, Yi Tay, Rishi Bommasani, Colin Raffel, Barret Zoph,
Sebastian Borgeaud, Dani Yogatama, et al. ``Emergent Abilities of Large
Language Models.'' arXiv, 2022. arXiv:2206.07682.

\mybibitem Westerwinter, Oliver. ``Transnational Public-Private Governance
Initiatives in World Politics: Introducing a New Dataset.'' \emph{The
Review of International Organizations} 16, no. 1 (2021): 137--74.
\url{https://doi.org/10.1007/s11558-019-09366-w}.

\mybibitem World Trade Organization. ``Agreement on the Application of Sanitary and
Phytosanitary Measures (SPS Agreement),'' n.d.
\url{https://www.wto.org/english/tratop_e/sps_e/spsagr_e.htm}.

\mybibitem World Trade Organization. ``Agreement on Trade-Related Aspects of
Intellectual Property Rights (TRIPS),'' n.d.
\url{https://www.wto.org/english/docs_e/legal_e/31bis_trips_01_e.htm}.

\mybibitem World Trade Organization. ``DS161: Korea --- Measures Affecting Imports
of Fresh, Chilled and Frozen Beef,'' January 10, 2001.
\url{https://www.wto.org/english/tratop_e/dispu_e/cases_e/ds161_e.htm}.

\mybibitem World Trade Organization. ``DS285: United States --- Measures Affecting
the Cross-Border Supply of Gambling and Betting Services.'' Accessed
August 18, 2023.
\url{https://www.wto.org/english/tratop_e/dispu_e/cases_e/ds285_e.htm}.

\mybibitem World Trade Organization. ``DS363: China --- Measures Affecting Trading
Rights and Distribution Services for Certain Publications and
Audiovisual Entertainment Products,'' December 21, 2019.
\url{https://www.wto.org/english/tratop_e/dispu_e/cases_e/ds363_e.htm}.

\mybibitem World Trade Organization. ``The General Agreement on Tariffs and Trade
(GATT),'' n.d.
\url{https://www.wto.org/english/docs_e/legal_e/gatt47_01_e.htm}.

\mybibitem World Trade Organization. ``General Agreement on Trade in Services
(GATS),'' n.d.
\url{https://www.wto.org/english/docs_e/legal_e/26-gats_01_e.htm}.

\mybibitem World Trade Organization. ``Understanding the WTO: Settling Disputes.''
Accessed May 23, 2023.
\url{https://www.wto.org/english/thewto_e/whatis_e/tif_e/disp1_e.htm}.

\mybibitem Ziegler, Andreas R., and David Sifonios. ``The Assessment of
Environmental Risks and the Regulation of Process and Production Methods
(PPMs) in International Trade Law.'' In \emph{Risk and the Regulation of
Uncertainty in International Law}, edited by Mónika Ambrus, Rosemary
Rayfuse, and Wouter Werner, 219--36. Oxford, UK: Oxford University
Press, 2017.

\end{document}